\documentclass{article} 
\usepackage{iclr2027_conference,times}

\usepackage{amsmath,amsfonts,bm}

\def\eqref#1{equation~\ref{#1}}

\def\1{\bm{1}}

\DeclareMathAlphabet{\mathsfit}{\encodingdefault}{\sfdefault}{m}{sl}
\SetMathAlphabet{\mathsfit}{bold}{\encodingdefault}{\sfdefault}{bx}{n}

\usepackage{titlesec}
\titlespacing*{\paragraph}
  {0pt}   
  {1ex}   
  {0.5em} 
  
\usepackage{hyperref}
\usepackage{url}
\usepackage{graphicx}
\usepackage{enumitem}
\usepackage{booktabs}
\usepackage{threeparttable}
\usepackage{xcolor}
\usepackage{subcaption}
\usepackage[table]{xcolor}
\usepackage{wrapfig}
\usepackage{multirow}
\usepackage{nicematrix}
\newcommand{\method}{\textsc{{SparseRepairer}}}
\usepackage[normalem]{ulem}
\usepackage{arydshln}
\usepackage{soul}
\usepackage{colortbl}
\usepackage{algorithm}
\usepackage{algpseudocode}

\usepackage{float}    
\usepackage{caption}  

\definecolor{hmblue}{RGB}{110,165,250}
\newcommand{\hmc}[1]{\cellcolor{hmblue!#1}}

\usepackage{booktabs}
\usepackage{diagbox}
\usepackage{xcolor}
\usepackage{colortbl}
\usepackage{enumitem}

\definecolor{baseblue}{HTML}{E6EFF6}
\definecolor{localblue}{HTML}{F0F4F7}
\definecolor{inheritrose}{HTML}{FAEEEB}
\definecolor{bothrose}{HTML}{F3DFDC}

\usepackage[noEnd=true,indLines=true]{algpseudocodex}
\usepackage{xcolor}
\definecolor{methodblue}{RGB}{31,92,163}

\title{Sparse Attention with\\ Entropy-Triggered KV Refresh}

\author{Yilin Guan \quad Sungwon Kim \quad Tianjian Li \quad Daniel Khashabi \\
Johns Hopkins University \\
{\small\texttt{yguan29@jh.edu} \quad \texttt{danielk@jhu.edu}}
}

\iclrfinalcopy 
\begin{document}

\maketitle

\newcommand{\ie}{\emph{i.e.,}}

\begin{abstract}

Sparse attention reduces the per-step decoding cost of language models trained with full attention. However, it can lower accuracy and lengthen generations, offsetting the per-step savings with additional tokens.
We find that sparse-attention errors are most detrimental at steps where the full-attention next-token entropy is high: applying sparse attention only at these steps degrades accuracy far more than applying it at an equal number of randomly selected steps. 
We further find that most of the degradation at these high-entropy steps stems from errors accumulated in the key-value (KV) cache during earlier sparse decoding, which persist even when the current step uses full attention. 
Motivated by these findings, we propose \method{}, a method which uses sparse next-token entropy to trigger a KV refresh: a full-attention pass that recomputes and replaces the cached KV states to yield the dense next-token distribution for sampling. 
Across three reasoning models and four benchmarks, \method{} improves accuracy by up to 18.0 percentage points and reduces generation length by up to 49.1\% over Quest, while achieving better accuracy--efficiency trade-offs than periodic refresh with both Quest and TidalDecode. With an optimized GPU implementation, \method{} matches dense decoding accuracy with 23.6\% lower wall-clock time and runs 7.9\% faster than periodic refresh on 87 long-generation MATH-500 problems.

\end{abstract}

\section{Introduction}

Modern language models thrive at solving hard problems by generating
\textit{very long} chains of reasoning~\citep{wei2022chain,snell2025scaling}, 
which has served as a linchpin for recent breakthroughs in  science~\citep{alman2026truly,riordan2026short}. 
Such reasoning traces can easily span thousands, if not millions, of tokens.
At these lengths, attention during decoding is bound by memory reads of the KV cache~\citep{shazeer2019fasttransformerdecodingwritehead, pope2022efficientlyscalingtransformerinference}: each step loads the KV cache of every preceding token, so the cumulative KV traffic of a trace grows quadratically with its length. Query-aware sparse attention~\citep{tang2024quest, xiao2024infllm, chen2025magicpig} reduces this cost by keeping the full cache in memory but attending to only a selected subset at each step. Because these methods apply to pretrained models at inference time, they offer a promising way to reduce the inference cost of long-form reasoning.

However, this reduction in KV-cache accesses can compromise generation quality.
It perturbs the attention outputs, alters the KV states propagated to later layers, and consequently affects the cache at future decoding steps and ultimately the next-token logits~\citep{sun2026rectified, yang2025tidaldecode}. Generation under sparse attention therefore drifts from the trajectory that full attention would produce. This drift is especially pronounced in long-form reasoning, where deviations compound across decoding steps~\citep{sun2026rectified}.
In practice, sparse decoding produces substantially longer generations and lower accuracy than dense decoding~\citep{gao2026sparse, hu-etal-2026-lil}. This undermines the very motivation for sparsity: \textit{lower generation quality results in longer reasoning traces which 
offset the computational savings at each decoding step.}

\begin{figure}[t]
    \centering
    \includegraphics[width=\linewidth]{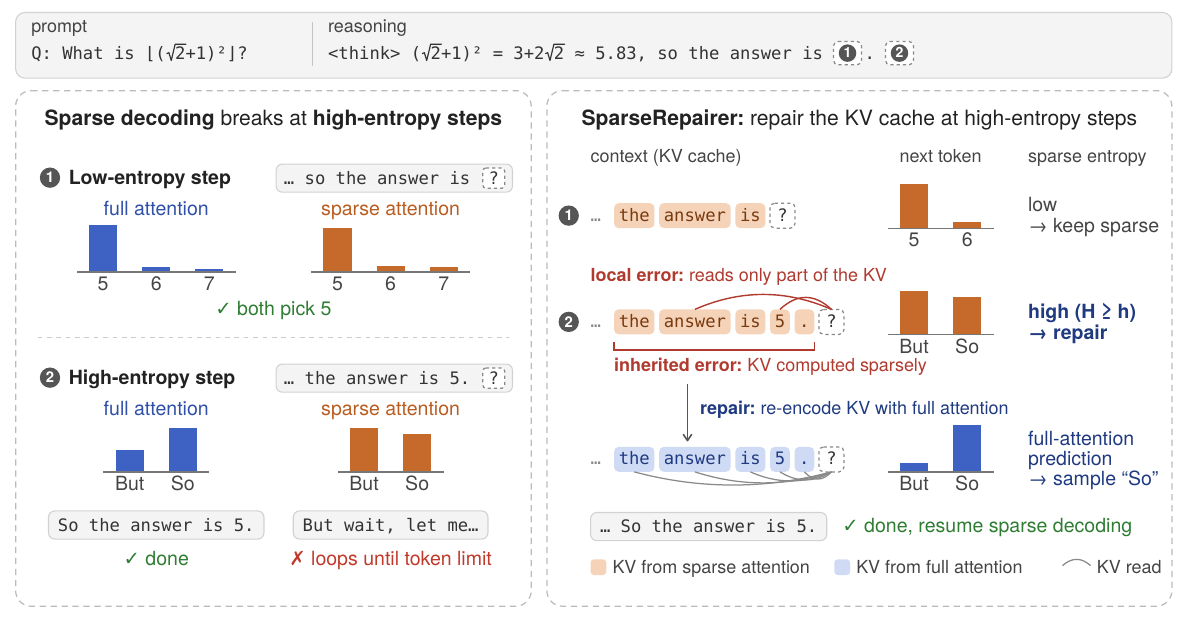}
    \caption{\textbf{Sparse decoding errors become most consequential at high-entropy steps.}
    \textbf{(Left)} Sparse and dense attention usually agree on the next token at low-entropy steps but diverge at high-entropy steps, potentially leading to prolonged reasoning instead of reaching a conclusion.
    \textbf{(Right)} \method{} identifies these important steps using sparse next-token entropy, refreshes the accumulated KV states with full attention, and samples the next token from the resulting dense distribution.
    }
    \label{fig:overview}
\end{figure}

We study the degradation of long-form reasoning under sparse attention, based on three key findings:

\begin{itemize}[leftmargin=*, itemsep=3pt]
    \item \textbf{Finding \#1:}
    To understand this drift, we ask at which decoding steps sparse and dense decoding diverge. 
    We compare them on identical prefixes, we observe that \emph{the discrepancy between their next-token distributions increases with the entropy of the dense prediction} (Sec~\ref{sec:tv}).

    \item \textbf{Finding \#2:}
    Larger per-step divergence, however, need not necessarily imply worse final answers.
    To determine whether these divergences affect final performance, we decode with full attention but switch to sparse attention only at steps whose dense entropy is high, and compare against switching at an equal fraction of randomly chosen steps (Sec~\ref{sec:entropy_sensitivity}).
    On Qwen3-8B and MATH-500, applying sparse attention at \textit{high-entropy steps} (approximately 15\% of decoding steps) reduces accuracy by 10.8 percentage points, compared with a 1.4-point accuracy drop when intervening at \textit{randomly selected} steps at the same rate (Table~\ref{tab:math500-entropy-sweep}).
    Thus, \emph{high-entropy steps are disproportionately sensitive to sparse attention}, as illustrated in Fig.~\ref{fig:overview} (left).

    \item \textbf{Finding \#3:}
    We identify two sources of error that contribute to the difference between sparse and dense predictions:
    (1) \emph{Local error} arises from attending to only a subset of past tokens at the current step.
    (2) \emph{Inherited error} arises from KV states perturbed by earlier sparse decoding steps.

    At the same high-entropy intervention steps, we isolate these effects by independently varying whether the current step uses sparse attention and whether its KV history was computed under sparse attention.
    We find that inherited error accounts for most of the degradation at high-entropy steps (Sec~\ref{sec:decomposition}).
    In Qwen3-8B on MATH-500, local error alone lowers accuracy by 1.2 points, whereas inherited error alone lowers it by 8.0 points.
    This suggests that \emph{restoring full attention for the current step alone is insufficient, as the dominant source of error already resides in the cache}.
\end{itemize}

These findings motivate \method{}, which \textit{selectively} refreshes accumulated KV errors at high-entropy decoding steps.
As shown in \autoref{fig:overview} (right),
at each step, \method~computes the next-token distribution using sparse attention and checks its entropy. When entropy exceeds a threshold, \method~re-encodes the KV states of tokens generated since the previous refresh using dense attention, updates the cache, and samples the next token from the resulting dense distribution. \method{} uses sparse next-token entropy as a proxy for the dense entropy (used in our initial analysis) allowing it to identify refresh steps without performing an additional dense forward pass.
Sec~\ref{sec:entropy_proxy} shows that this proxy is effective, with an AUROC of 0.94--0.97 across thresholds. 

We evaluate \method{} on three reasoning models and four benchmarks using Quest~\citep{tang2024quest} as the sparse-attention backend, and further validate its compatibility with TidalDecode~\citep{yang2025tidaldecode}. \method{} improves accuracy by up to 18.0 percentage points and reduces generation length by up to 49.1\% over Quest, while achieving favorable accuracy--efficiency trade-offs against periodic refresh with both sparse attention methods. With an optimized GPU implementation, \method{} also matches dense decoding accuracy with 23.6\% lower wall-clock time and runs 7.9\% faster than periodic refresh on 87 MATH-500 problems with dense-decoding trajectories exceeding 8K tokens.

Our contributions are as follows:
\begin{itemize}[leftmargin=*]
    \item We identify that sparse-attention errors are most detrimental at steps where the full-attention next token entropy is high. Applying sparse attention only at such steps severely degrades performance.
    \item We isolate \emph{local} and \emph{inherited} error at these steps and demonstrate that inherited error, the error from KV cache from past sparse decoding steps, accounts for most of the degradation.
    \item We propose \method{}, which uses sparse next-token entropy to trigger KV refresh. At comparable refresh frequencies, \method{} generally achieves higher accuracy with shorter generations than periodic refresh. We further demonstrate its practical efficiency through an optimized GPU implementation, achieving wall-clock time reductions over both dense decoding and periodic refresh.

\end{itemize}

\section{Background and Related Work}


\paragraph{Cache drift under sparse decoding.} 
Prior work has observed that sparse decoding accumulates error in the KV cache and degrades long generations~\citep{yang2025tidaldecode, sun2026rectified, yang2026less}.
Because attention outputs determine the keys and values of subsequent layers, the resulting distributional shift propagates across layers and into subsequent decoding steps~\citep{willette2025delta}.
In reasoning models, this accumulated error also lengthens generation, potentially offsetting the per-step computational savings of sparse attention~\citep{gao2026sparse, hu-etal-2026-lil}.
ReSA~\citep{sun2026rectified} \textit{periodically} refreshes the accumulated KV cache with dense re-encoding. 
RetroAttention~\citep{ICLR2026_f4daa773} retrospectively supplements past attention outputs with KV pages selected at later steps.
Rather than repairing at fixed intervals, \method{} instead identifies \textit{when} accumulated KV errors are most consequential and triggers dense history refresh only at these high-entropy steps.


\paragraph{High-entropy tokens and generation.} Prior work has explored high-entropy tokens for allocating computation during training and inference. \citet{wang2025beyond} show that restricting policy updates to a minority of high-entropy tokens preserves or improves reasoning performance. \citet{chen2026rstitch} route high uncertainty tokens to a larger model in a speculative decoding setup.  
Other works use predictive entropy to choose branching points during reasoning \citep{xu2025dts, wang-etal-2025-cautious, zhou2026reasoningsamplingcuttingdecision, scalena2025eager}. \citet{kai2026informationawarekvcachecompression} use entropy as a heuristic for KV cache compression, and DiffAdapt~\citep{liu2026diffadapt} uses reasoning-trace entropy to choose an inference strategy.  
While these results signal the importance of high entropy tokens, \method{} uses them specifically to decide when accumulated KV-cache errors warrant refresh.

\section{Where Sparse Attention Breaks Reasoning}
\label{sec:mechanism}
During autoregressive decoding, query-aware sparse attention reduces KV-cache memory traffic and attention computation by attending to only a selected subset of previous KV states at each step. The sparsity perturbs the attention outputs and, consequently, the hidden representations of the current token. These changes propagate to the KV states computed at subsequent layers and persist in the cache for future decoding steps. We distinguish two sources of error in sparse attention: \emph{local error}, caused by attending to only a subset of cached KV states at the current step, and \emph{inherited error}, caused by KV states computed under sparse attention at earlier steps.

We first identify the decoding steps where sparse predictions differ most from dense predictions (Sec.~\ref{sec:tv}). We then test whether sparsity at these steps degrades reasoning performance (Sec.~\ref{sec:entropy_sensitivity}) and isolate the contributions of local and inherited errors (Sec.~\ref{sec:decomposition}).


\subsection{Where Sparse Predictions Depart from Dense Decoding}
\label{sec:tv}


\paragraph{Motivation.} To understand why sparse attention lengthens reasoning traces and lowers accuracy, 
we identify the decoding steps at which sparse predictions
depart from dense ones. Instead of comparing freely decoded sparse and dense trajectories, we fix a dense trajectory and 
branch at each decoding step to recompute the next-token distribution under sparse attention.
Both branches use the same token prefix, but the sparse branch recomputes its KV history and next-token prediction under sparse attention.

\begin{wrapfigure}[18]{r}{0.43\linewidth}
    \centering
    \vspace{-1.9em}
    \includegraphics[width=1\linewidth]{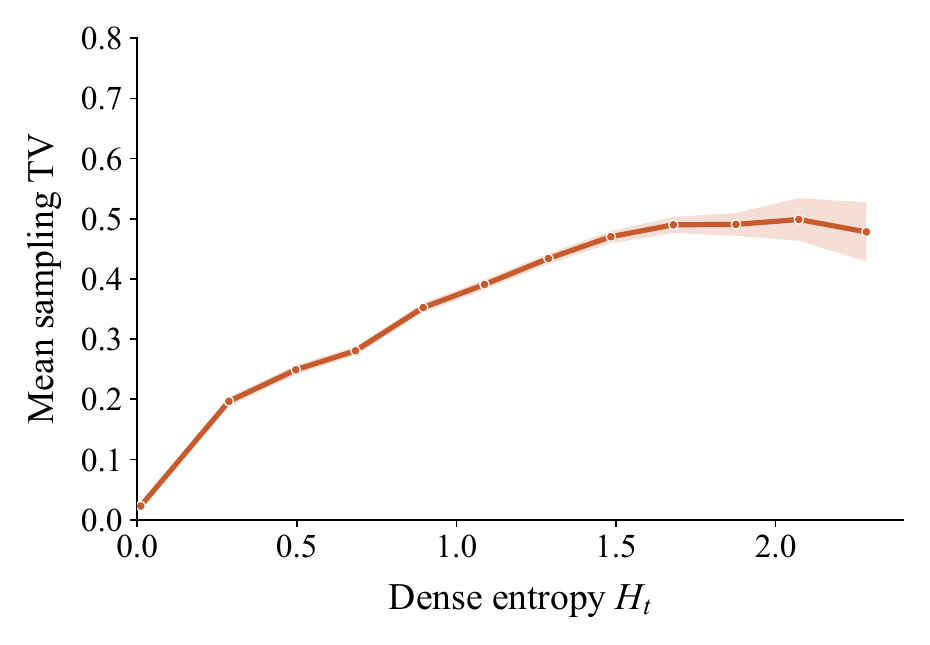}
    \vspace{-1.9em}
    \caption{
            \textbf{Sparse and dense next-token distributions diverge more at high-entropy steps} (\S\ref{sec:tv}). 
            Along fixed dense-decoded trajectories, we plot the mean total variation (TV) distance between sparse and dense next-token distributions against dense next-token entropy.
            Shading indicates 95\% confidence intervals.
    }
    \label{fig:sampling_tv}
    \vspace{-0.6em}
\end{wrapfigure}
\paragraph{Setup.} Dense trajectories are generated by Qwen3-8B on MATH-500. Both branches encode the prompt with full attention. 
For each fixed token prefix, the dense branch keeps the original dense KV cache, while the sparse branch recomputes the KV cache with Quest and attends to it sparsely. Let $p^D_t$ and $p^S_t$ denote the next-token distributions of the dense and sparse branches at temperature 0.6 before top-$p$ truncation, and let $H_t = -\sum_v p^D_t(v)\log p^D_t(v)$ denote the entropy of the dense branch at decode step $t$, over the vocabulary $v$. Applying top-$p$ truncation and renormalization to each distribution yields the actual sampling distributions $q^D_t$ and $q^S_t$. We measure their discrepancy by total variation distance, $\mathrm{TV}(q^D_t,q^S_t) = \tfrac{1}{2}\sum_v |q^D_t(v)-q^S_t(v)|$. $\mathrm{TV}$ is the fraction of probability mass on which the sparse and dense distributions disagree. It ranges from 0 when the two distributions are identical to 1 when their supports are disjoint.

\paragraph{Findings.} Figure~\ref{fig:sampling_tv} shows that mean sampling $\mathrm{TV}$ increases with dense next-token entropy. Even with the same prefix, sparse attention induces larger changes in the sampling distribution at positions where dense decoding is more uncertain. This suggests that 
next-token entropy is a useful indicator of sensitivity to sparse decoding. \textbf{Thus, sparse and dense predictions diverge most at high-entropy steps and remain substantially closer at low-entropy steps.}

\subsection{Sparse Attention at High-Entropy Steps Degrades Reasoning}
\label{sec:entropy_sensitivity}

\paragraph{Motivation.} Sec.~\ref{sec:tv} shows that sparse and dense predictions diverge most at high-entropy steps. However, whether these divergences degrade downstream reasoning performance remains unclear.
We therefore test whether 
applying sparse attention at high-entropy steps degrades reasoning more than applying it at randomly selected steps.


\paragraph{Setup.} We use Qwen3-8B on MATH-500 with Quest~\citep{tang2024quest} at a token budget of 448 and a 12{,}288-token generation limit. We decode with full attention and switch to sparse attention only at selected steps to measure their sensitivity.
We call the switch to sparse attention an \emph{intervention}. 
At an intervened step $t$, 
we compute the next-token distribution that sparse decoding would produce for the same token prefix.
To obtain this distribution, we recompute the generated prefix's KV states under sparse attention, attend sparsely to these states at step $t$, and sample the next token from the resulting distribution.
The prediction therefore carries both local and inherited error. We then discard the recomputed states and continue with full attention from the original dense cache, so the only thing an intervention changes for later steps is the token $x_t$. The prompt is always encoded with full attention. We select the intervened steps in two ways. Let $H_t$ be the entropy of the full-attention next-token distribution at step $t$. The high-entropy policy intervenes whenever $H_t \geq \tau$, and the random policy intervenes at randomly chosen steps with the same total number of intervened steps.

\begin{figure}[h]
      \centering
      \begin{subfigure}[t]{0.48\linewidth}
          \centering
          \includegraphics[width=\linewidth]{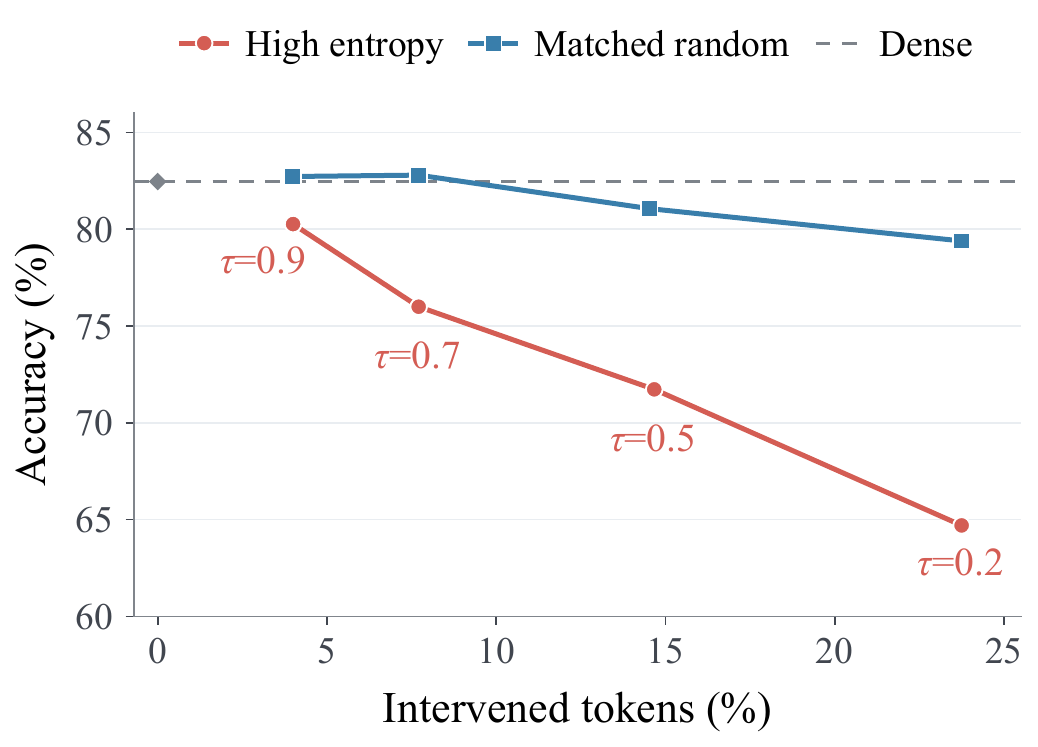}
          \label{fig:accuracy_rate}
      \end{subfigure}
      \hfill
      \begin{subfigure}[t]{0.48\linewidth}
          \centering
          \includegraphics[width=\linewidth]{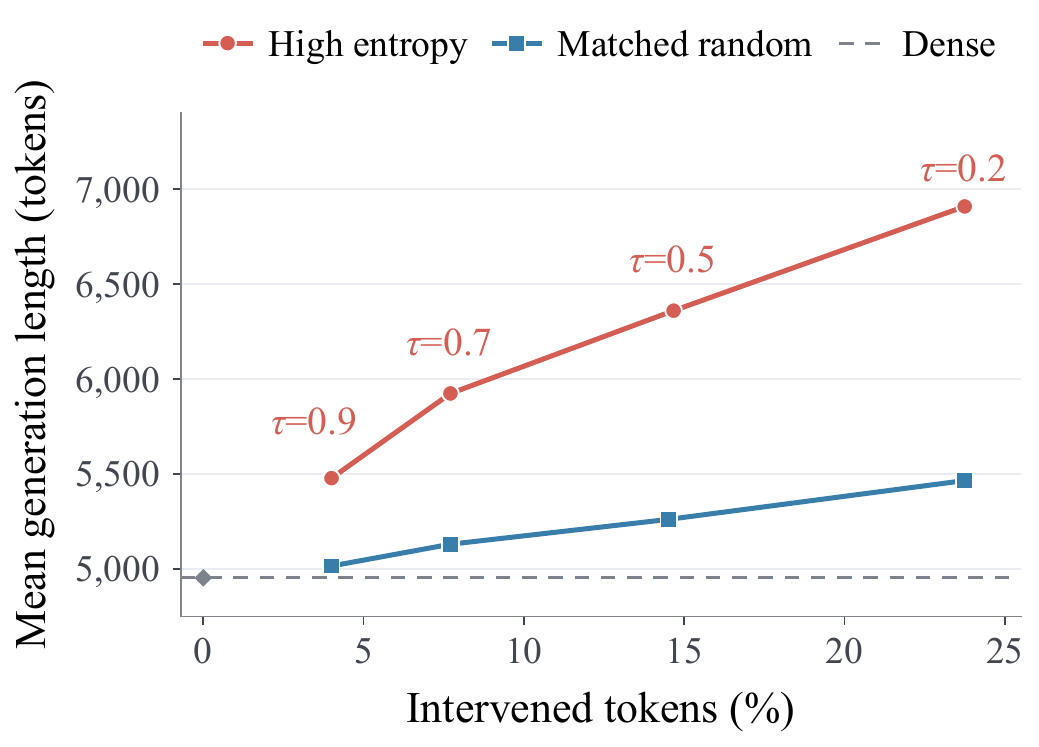}
          \label{fig:length_rate}
      \end{subfigure}
      \vspace{-0.3cm}
      \caption{\textbf{Sparse-attention interventions at high-entropy steps lower accuracy (left) and lengthen generations (right) 
      far more than rate-matched random interventions}      
      (\S\ref{sec:entropy_sensitivity}). Each point on the high-entropy curve corresponds to a threshold $\tau$, and the random policy is matched to that rate. The dashed line is dense decoding. Exact numbers are reported in Table~\ref{tab:math500-entropy-sweep} (Appendix~\ref{app:intervention}).
      }
      \label{fig:motivation_rate}
  \end{figure}
\paragraph{Findings.} 
Figure~\ref{fig:motivation_rate} shows that high-entropy interventions cause substantially greater degradation than rate-matched random interventions.
When intervening at steps with $H_t \geq 0.5$ (14.7\% of decoding steps), accuracy drops from 82.5\% to 71.7\%, the average generation length grows from 5.0k to 6.4k tokens, 
and the truncation rate rises from
8.1\% to 22.5\%. 
In contrast, rate-matched random interventions yield 81.1\% accuracy, 5.3k generated tokens, and a 10.3\% truncation rate, \textit{all close to dense decoding}.
We report detailed results in \autoref{tab:math500-entropy-sweep}.

The gap holds at every threshold and widens as $\tau$ is lowered. 
As $\tau$ decreases, the high-entropy policy intervenes at more steps (4.0\% to 23.8\%), reducing accuracy from 80.3\% to 64.7\% and increasing mean generation length from 5.5k to 6.9k tokens.
Random interventions at the same rates stay close to dense decoding, losing at most 3.1 points of accuracy even when applied to 23.7\% of steps. 
We observe a similar pattern on GPQA-Diamond: high-entropy interventions substantially lengthen generations, whereas approximately rate-matched random interventions have little effect
(\ref{app:gpqa-intervention}).

These results show that \textbf{high-entropy steps are disproportionately sensitive to sparse attention}: 
applying sparsity at the same rate can have substantially different effects on reasoning performance depending on which steps are selected.

\subsection{Tracing the degradation to sparse KV history}
\label{sec:decomposition}

\paragraph{Motivation.} 
Sparse decoding introduces two sources of prediction error: \textit{local error} from attending to only a subset of past KV states, and \textit{inherited error} from KV states computed under sparse attention at earlier steps.
To understand how these perturbations turn into performance degradation, we isolate the two types of error at high-entropy steps ($H_t \ge 0.5$).

\paragraph{Setup.} 
To isolate inherited error, we recompute the generated prefix's KV cache under \textit{sparse} attention but use \textit{full  attention} for the current prediction.
To isolate local error, we retain the dense KV states, and at intervention steps, the model sparsely attends to past KV states. Both interventions follow the protocol of Sec.~\ref{sec:entropy_sensitivity}: after sampling, the altered KV states are discarded and decoding continues from the dense cache. All non-intervened steps use full attention. Together with dense decoding and the combined intervention, these yield the four configurations in Table~\ref{tab:decomposition}.

\newcommand{\dl}[1]{{\scriptsize\textcolor{gray}{($#1$)}}}

\begin{table}[h!]
    \centering
    \small
    \setlength{\tabcolsep}{5pt}
    \definecolor{localfg}{HTML}{8E3B6F}
    \definecolor{inhfg}{HTML}{B0502A}
    \colorlet{localbg}{inhfg!3}
    \colorlet{inhbg}{inhfg!15}
    \colorlet{bothbg}{inhfg!21}
    \begin{tabular}{l r@{}l r r@{}l r}
        \toprule
        & \multicolumn{6}{c}{\textbf{Attention at intervened steps} ($H_t \geq 0.5$)} \\
        \cmidrule(lr){2-7}
        & \multicolumn{3}{c}{\textbf{Dense}} & \multicolumn{3}{c}{\textbf{Sparse} \scriptsize\textcolor{localfg}{(+ local error)}} \\
        \cmidrule(lr){2-4} \cmidrule(lr){5-7}
        \textbf{History KV states}\,$\downarrow$ & \multicolumn{2}{c}{Acc.\ (\%)\,$\uparrow$} & Tokens\,$\downarrow$ & \multicolumn{2}{c}{Acc.\ (\%)\,$\uparrow$} & Tokens\,$\downarrow$ \\
        \midrule
        Dense & \textbf{82.5} & & \textbf{4,953} & \cellcolor{localbg}81.3 & \cellcolor{localbg}\hspace{3pt}\dl{-1.2} & \cellcolor{localbg}5,161 \\
        & \multicolumn{3}{c}{\scriptsize\textcolor{gray}{no error (dense decoding)}} & \multicolumn{3}{c}{\cellcolor{localbg}\scriptsize\textcolor{localfg}{local error only}} \\
        \midrule
        Sparse & \cellcolor{inhbg}74.5 & \cellcolor{inhbg}\hspace{3pt}\dl{-8.0} & \cellcolor{inhbg}6,168 & \cellcolor{bothbg}71.7 & \cellcolor{bothbg}\hspace{3pt}\dl{-10.8} & \cellcolor{bothbg}6,360 \\
        \scriptsize\textcolor{inhfg}{(+ inherited error)} & \multicolumn{3}{c}{\cellcolor{inhbg}\scriptsize\textcolor{inhfg}{inherited error only}} & \multicolumn{3}{c}{\cellcolor{bothbg}\scriptsize\textcolor{localfg}{local} + \textcolor{inhfg}{inherited error}} \\
        \bottomrule
    \end{tabular}
    \caption{\textbf{Inherited error accounts for most of the degradation at high-entropy steps} (\S\ref{sec:decomposition}). Results on Qwen3-8B and MATH-500. Columns vary the attention pattern at intervened steps ($H_t \geq 0.5$), and rows vary how the KV history was computed. 
    KV history computed under sparse attention introduces inherited error
    (orange), 
    while attending sparsely to past KV states at intervention steps introduces local error
    (purple); 
    darker shading indicates a larger accuracy drop. Gray values show accuracy changes relative to dense decoding and dense KV history. 
    }
    \label{tab:decomposition}
\end{table}

\paragraph{Findings.} Table~\ref{tab:decomposition} shows that inherited error alone reduces accuracy by 8.0 points (82.5\% to 74.5\%) and lengthens generations from 5.0k to 6.2k tokens. In contrast, local error alone costs just 1.2 points (81.3\%), with little change in generation length (5.2k tokens). 

These results show that \textbf{errors inherited from previously computed KV states account for most of the degradation}.
Using dense attention only at the current step is therefore not enough, as it recovers only 2.8 of the 10.8 lost points (71.7\% to 74.5\%), because the errors it would need to fix are inherited in the cache.

\section{\method{}: Entropy-triggered KV Cache Repair}
Section~\ref{sec:mechanism} shows that high-entropy steps are sensitive to sparse attention, with most of the degradation arising from inherited errors that persist even when the current step uses full attention.
These findings motivate \method{}, which adaptively refreshes the KV history at high-entropy steps and samples from the resulting dense distribution, rather than refreshing at fixed intervals.



\begin{figure}[h!]
    \centering
    \includegraphics[width=\linewidth, 
    ]{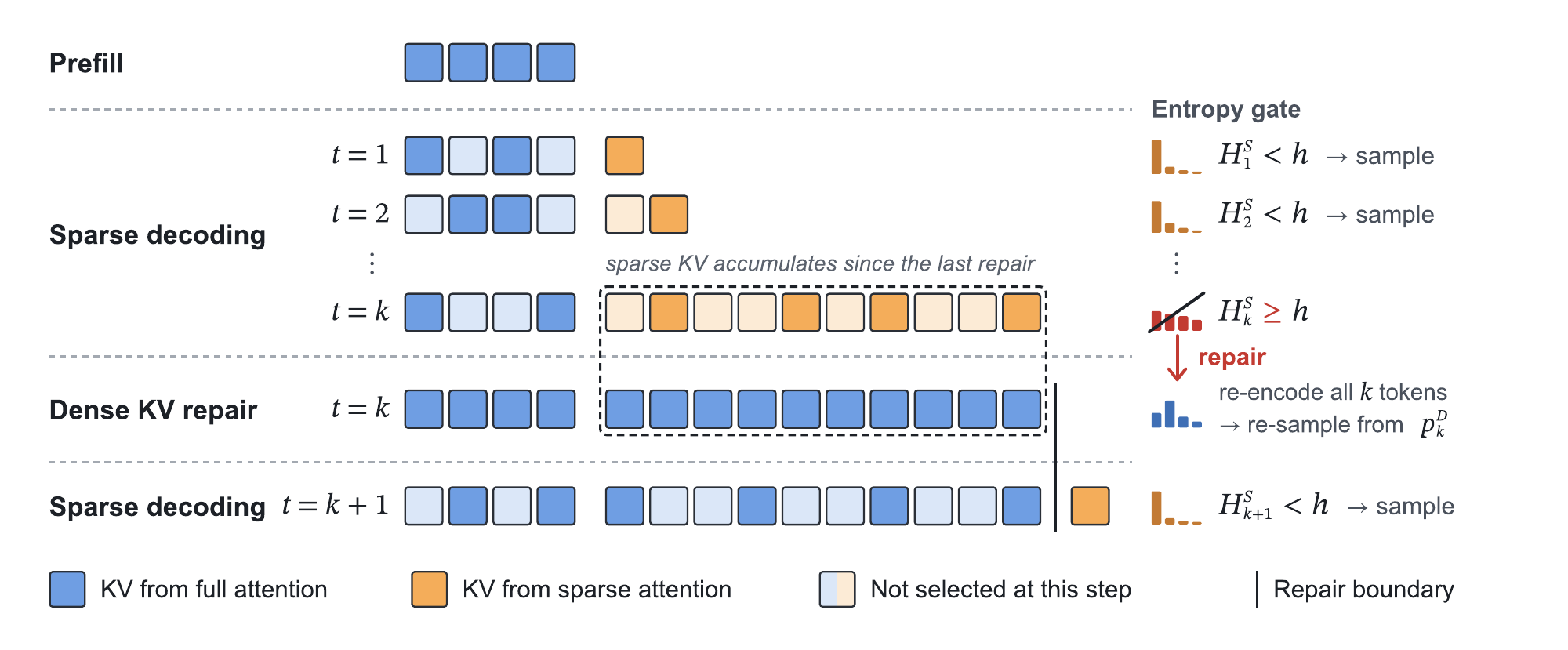}
    \caption{
    \textbf{Token-level workflow of \method.}
    \method{} decodes with sparse attention by default.
    When sparse next-token entropy exceeds the threshold, it re-encodes the tokens generated since the previous refresh boundary with full attention, replaces their KV states, and samples from the resulting dense next-token distribution before resuming sparse decoding.
    }
    \label{fig:sparse_repairer_mechanism}
\end{figure}

\subsection{High-entropy Triggered KV Cache Repair}
\label{sec:sparse_repairer_method}

Figure~\ref{fig:sparse_repairer_mechanism} shows \method's workflow. It starts with a standard dense prefill over the prompt. 
At each decoding step, \method first uses sparse attention to compute the next-token distribution.
Let $H_t^S$ denote the entropy of the sparse next-token distribution at step $t$. If $H_t^S < h$, \method~samples the next token from this sparse distribution and proceeds to the next sparse decoding step. As generation continues in this way, the KV states of newly generated tokens are computed under sparse attention and accumulate in the cache.

When $H_t^S \ge h$, \method~pauses sparse decoding and refreshes the KV states accumulated since the previous refresh. 
Specifically, it re-encodes these tokens with full causal attention, using the previously refreshed KV history as context, and replaces their sparse KV states with the recomputed dense ones.
The replay also yields the dense next-token distribution at step $t$, from which \method~samples the next token. Sparse decoding then resumes from the repaired cache.

To avoid triggering multiple refreshes in close succession, we require at least $m$ generated tokens between consecutive refreshes. We use $m=8$ in our experiments. Alg~\ref{alg:sparserepairer} illustrates the procedure.

\begin{algorithm}[h!]
\caption{\method: Entropy-Triggered KV Refresh}
\label{alg:sparserepairer}
\begin{algorithmic}[1]
\Require Prompt $P$, sparse backend $\mathcal{S}$,
entropy threshold $h$, minimum refresh interval $m$

\State $(K,p) \gets \Call{DensePrefill}{P}; \quad r \gets 0$

\For{$t = 1, 2, \ldots$}
    \State $x_t \sim p$
    \State \textbf{if} $x_t = \mathrm{EOS}$ \textbf{then break}
    \State $(K,p) \gets \Call{SparseStep}{x_t,K,\mathcal{S}}$

    \If{\textcolor{methodblue}{$H(p) \geq h$}
        \textbf{and} $t-r \geq m$}
        \State $(K_{r+1:t},p) \gets
        \textcolor{methodblue}{
        \Call{DenseRefresh}{x_{r+1:t},K_{\leq r}}}$
        \State $r \gets t$
    \EndIf
\EndFor
\end{algorithmic}
\end{algorithm}






Here, $K_{\leq 0}$ denotes the dense prompt cache, and
\textsc{DenseRefresh} replaces the KV states generated since
the previous refresh while returning the updated next-token
distribution.
\textcolor{methodblue}{Blue} highlights the two key operations
of \method: \textcolor{methodblue}{the entropy-based
refresh trigger and the dense KV refresh}.

\subsection{Sparse Entropy Is a Faithful Proxy for Dense Uncertainty}
\label{sec:entropy_proxy}

\begin{wrapfigure}{r}{0.45\linewidth}
    \centering
    \vspace{-0.9em}
    \includegraphics[width=0.9\linewidth]{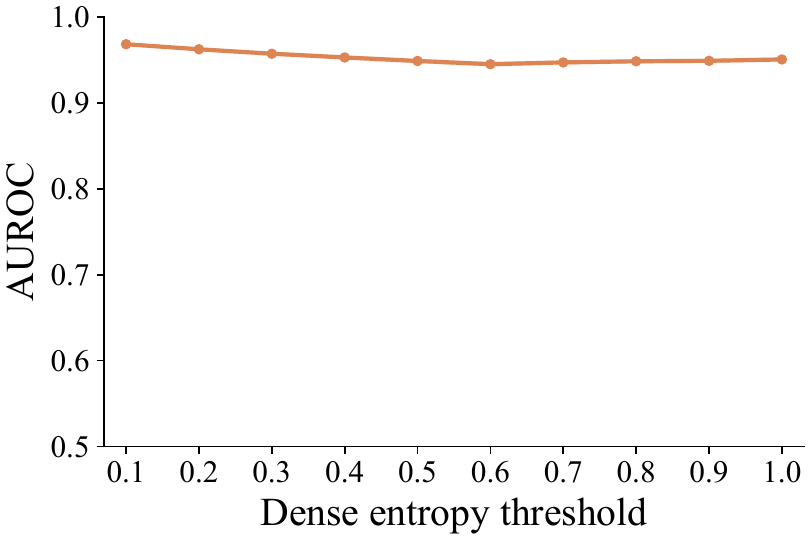}
    \caption{
        AUROC of sparse entropy for identifying steps with dense entropy $\geq \tau$,
        across different thresholds $\tau$ on Qwen3-8B and MATH-500.
        \textbf{Sparse entropy consistently identifies decoding steps with high dense entropy.}
    }
    \label{fig:auroc}
\end{wrapfigure}

To evaluate whether sparse entropy reliably identifies high-dense-entropy steps, we use paired dense and sparse predictions from the fixed-prefix replay in Sec.~\ref{sec:tv}. Let $H_t^D$ and $H_t^S$ denote their next-token entropies, computed at temperature 0.6 before top-$p$ truncation. For each dense-entropy threshold $\tau$, we label steps with $H_t^D \geq \tau$ as high-entropy steps and evaluate how well $H_t^S$ identifies them using the area under the receiver operating characteristic curve (AUROC). We vary $\tau$ from 0.1 to 1.0.

Figure~\ref{fig:auroc} shows that sparse entropy consistently identifies decoding steps with high dense entropy, with AUROCs of approximately 0.94--0.97 across the tested thresholds. We therefore use sparse next-token entropy in \method~to decide when to refresh the KV cache.

\section{Experiments}
\label{sec:experiments}

\subsection{Experimental Setup}
\label{sec:experimental_setup}

\paragraph{Models and benchmarks.}
We evaluate three reasoning models: Qwen3-8B and Qwen3-4B in thinking mode \citep{yang2025qwen3technicalreport}, and DeepSeek-R1-Distill-Llama-8B \citep{guo2025deepseekr1}. We use four benchmarks: MATH-500 (500 mathematical problems; \citet{lightman2023lets}), AIME 2025 (30 competition problems; \citet{aops_aime_2025}), GPQA-Diamond (198 multiple-choice science questions; \citet{rein2023gpqagraduatelevelgoogleproofqa}), and LiveCodeBench v6 (175 code generation problems; \citet{jain2024livecodebench}).

\paragraph{Decoding settings.}
All methods use dense prefill, temperature 0.6, and top-$p$ 0.95.
For controlled comparison, Quest, ReSA-$k$, and \method{} use the same Quest-style sparse-attention backend with a page size of 64 tokens and per-query-head page selection.
Rather than using ReSA's GQA-grouped implementation, we instantiate its periodic rectification policy on this common backend, so that ReSA and \method{} differ primarily in when dense KV refresh is triggered. We set the attention-read budget of each benchmark to roughly 10\% of the dense model's mean generation length, which gives 448, 2,048, 1,024, and 1,024 tokens for MATH-500, AIME 2025, GPQA-Diamond, and LiveCodeBench, respectively. The maximum generation length is 32,768 tokens.

\paragraph{Compared Methods.}
\textbf{(1) Dense} uses full attention throughout decoding.
\textbf{(2) Quest} selects KV pages according to query-dependent scores computed from page-level key extrema \citep{tang2024quest}.
\textbf{(3) ReSA-$k$} ($k \in \{16,32\}$) applies periodic KV rectification every $k$ generated-token forward steps, using the same sparse backend and attention budget as Quest \citep{tang2024quest,sun2026rectified}.
\textbf{(4) \method{}} uses the same sparse backend and attention budget as ReSA. We treat the entropy threshold $h$ as a compute-control parameter rather than a performance-tuned hyperparameter. For comparison with ReSA-\(k\), we estimate candidate thresholds from the entropy distribution of Quest baseline trajectories and choose values that approximately match its refresh frequency. This procedure uses neither correctness labels nor \method{} accuracy (see Appendix~\ref{app:choosing_h}).

\paragraph{Metrics.}
We report accuracy (Acc.), mean generated tokens (Tokens), mean dense refresh calls (Repairs), and estimated decoding KV-memory access (KV Access). Truncation rates are reported in Appendix~\ref{app:truncation}.

Since decoding phase is memory bound, we primarily evaluate efficiency through estimated memory traffic, which consists of two parts: (1) \emph{KV-cache loading cost}: KV Access estimates the aggregate KV-cache entries accessed during decoding, including dense KV refreshes, normalized by the corresponding dense-decoding cost (Appendix~\ref{app:memory-access}). (2) \emph{Model-weight loading cost}: Tokens and Repairs serve as proxies for the cost of loading model weights during generation and refresh.
Lower values of Tokens, Repairs, and KV Access indicate better efficiency. 

\subsection{Main Results}
\label{sec:main_results}

\begin{table*}[h!]
  \centering
  \caption{Qwen3-8B results across four reasoning benchmarks. Parenthesized values beside \method{} denote the entropy threshold $h$. Tokens and Repairs are means per trajectory, and the parenthesized percentages in \method{} rows indicate changes relative to the ReSA configuration immediately above. Accuracy is a percentage; KV Access is normalized by Dense within each model--benchmark setting. Dashed lines separate the dense baseline from sparse methods.}
  \label{tab:qwen3_8b_main_results}
  \begingroup
  \small
  \setlength{\tabcolsep}{2pt}
  \renewcommand{\arraystretch}{1.12}

  \colorlet{qwenDelta}{green!60!black}
  \colorlet{qwenDeltaUp}{blue!75!black}

  \newcommand{\qwenChange}[2]{%
    #1\,\textcolor{qwenDelta}{\scriptsize(\ensuremath{#2}\%)}}
  \newcommand{\qwenChangeUp}[2]{%
    #1\,\textcolor{qwenDeltaUp}{\scriptsize(\ensuremath{#2}\%)}}
  \newcommand{\qwenName}[1]{%
    \textbf{\method}\,{\scriptsize
    \textcolor{black!60}{(#1)}}}

  \resizebox{\linewidth}{!}{%
  \begin{tabular}{@{}lllll@{\hspace{12pt}}lllll@{}}
  \toprule
  \multicolumn{5}{c}{\textbf{MATH-500} {\normalfont (500 questions)}}
  & \multicolumn{5}{c}{\textbf{AIME 2025} {\normalfont (30 questions)}} \\
  \cmidrule(lr){1-5}\cmidrule(lr){6-10}
  Method
  & Acc.$\uparrow$ & Tokens$\downarrow$ & Repairs$\downarrow$ & KV Access$\downarrow$
  & Method
  & Acc.$\uparrow$ & Tokens$\downarrow$ & Repairs$\downarrow$ & KV Access$\downarrow$ \\
  \midrule
  Dense
  & 86.3 & 5,413 & 0.0 & 1.000$\times$
  & Dense
  & 71.1 & 17,501 & 0.0 & 1.000$\times$ \\
  \addlinespace[1.5pt]
  \cdashline{1-5}[1pt/1.5pt]\cdashline{6-10}[1pt/1.5pt]
  \addlinespace[2pt]
  Quest
  & 67.0 & 13,274 & 0.0 & 1.176$\times$
  & Quest
  & 66.7 & 19,679 & 0.0 & 0.822$\times$ \\
  \addlinespace[4pt]
  ReSA-16
  & 81.0 & 8,149 & 508.8 & 0.822$\times$
  & ReSA-16
  & 65.6 & 18,080 & 1,129.4 & 0.813$\times$ \\
  \qwenName{0.5}
  & \textbf{83.9}
  & \qwenChange{6,932}{-14.9}
  & \qwenChange{392.9}{-22.8}
  & \qwenChange{0.633$\times$}{-23.0}
  & \qwenName{0.5}
  & 67.8
  & \qwenChange{17,523}{-3.1}
  & \qwenChange{995.4}{-11.9}
  & \qwenChange{0.772$\times$}{-5.0} \\
  \addlinespace[4pt]
  ReSA-32
  & 77.8 & 9,632 & 300.4 & 0.918$\times$
  & ReSA-32
  & 64.4 & 18,796 & 586.7 & 0.815$\times$ \\
  \qwenName{0.9}
  & 80.9
  & \qwenChange{8,246}{-14.4}
  & \qwenChange{230.0}{-23.4}
  & \qwenChange{0.747$\times$}{-18.6}
  & \qwenName{0.9}
  & \textbf{68.9}
  & \qwenChange{17,741}{-5.6}
  & \qwenChange{485.4}{-17.3}
  & \qwenChange{0.756$\times$}{-7.2} \\
  \midrule
  \multicolumn{5}{c}{\textbf{GPQA-Diamond} {\normalfont (198 questions)}}
  & \multicolumn{5}{c}{\textbf{LiveCodeBench v6} {\normalfont (175 questions)}} \\
  \cmidrule(lr){1-5}\cmidrule(lr){6-10}
  Method
  & Acc.$\uparrow$ & Tokens$\downarrow$ & Repairs$\downarrow$ & KV Access$\downarrow$
  & Method
  & Acc.$\uparrow$ & Tokens$\downarrow$ & Repairs$\downarrow$ & KV Access$\downarrow$ \\
  \midrule
  Dense
  & 62.6 & 9,776 & 0.0 & 1.000$\times$
  & Dense
  & 49.4 & 14,525 & 0.0 & 1.000$\times$ \\
  \addlinespace[1.5pt]
  \cdashline{1-5}[1pt/1.5pt]\cdashline{6-10}[1pt/1.5pt]
  \addlinespace[2pt]
  Quest
  & 54.3 & 15,963 & 0.0 & 1.154$\times$
  & Quest
  & 39.4 & 22,040 & 0.0 & 0.691$\times$ \\
  \addlinespace[4pt]
  ReSA-16
  & 60.4 & 11,008 & 687.5 & 0.822$\times$
  & ReSA-16
  & \textbf{49.7} & 17,157 & 1,071.7 & 0.605$\times$ \\
  \qwenName{0.5}
  & \textbf{61.9}
  & \qwenChange{10,351}{-6.0}
  & \qwenChangeUp{774.7}{+12.7}
  & \qwenChange{0.770$\times$}{-6.3}
  & \qwenName{0.9}
  & \textbf{49.7}
  & \qwenChange{16,870}{-1.7}
  & \qwenChange{762.8}{-28.8}
  & \qwenChange{0.566$\times$}{-6.4} \\
  \addlinespace[4pt]
  ReSA-32
  & 60.4 & 11,291 & 352.3 & 0.804$\times$
  & ReSA-32
  & 48.9 & 18,166 & 567.0 & 0.604$\times$ \\
  \qwenName{1.1}
  & 61.1
  & \qwenChange{11,226}{-0.6}
  & \qwenChange{311.6}{-11.6}
  & \qwenChange{0.795$\times$}{-1.1}
  & \qwenName{1.05}
  & \textbf{49.7}
  & \qwenChange{17,562}{-3.3}
  & \qwenChangeUp{572.5}{+1.0}
  & \qwenChange{0.576$\times$}{-4.6} \\
  \bottomrule
  \end{tabular}%
  }
  \endgroup
\end{table*}

\begin{table*}[h!]
  \centering
  \caption{Qwen3-4B results across four reasoning benchmarks.
  Notation follows Table~\ref{tab:qwen3_8b_main_results}.}
  \label{tab:qwen3_4b_main_results}
  \begingroup
  \small
  \setlength{\tabcolsep}{2pt}
  \renewcommand{\arraystretch}{1.12}

  \colorlet{qfourDelta}{green!60!black}
  \colorlet{qfourDeltaUp}{blue!75!black}

  \newcommand{\qfourChange}[2]{%
    #1\,\textcolor{qfourDelta}{\scriptsize(\ensuremath{#2}\%)}}
  \newcommand{\qfourChangeUp}[2]{%
    #1\,\textcolor{qfourDeltaUp}{\scriptsize(\ensuremath{#2}\%)}}
  \newcommand{\qfourName}[1]{%
    \textbf{\method{}}\,{\scriptsize
    \textcolor{black!60}{(#1)}}}

  \resizebox{\linewidth}{!}{%
  \begin{tabular}{@{}lllll@{\hspace{12pt}}lllll@{}}
  \toprule
  \multicolumn{5}{c}{\textbf{MATH-500} {\normalfont (500 questions)}}
  & \multicolumn{5}{c}{\textbf{AIME 2025} {\normalfont (30 questions)}} \\
  \cmidrule(lr){1-5}\cmidrule(lr){6-10}
  Method
  & Acc.$\uparrow$ & Tokens$\downarrow$ & Repairs$\downarrow$ & KV Access$\downarrow$
  & Method
  & Acc.$\uparrow$ & Tokens$\downarrow$ & Repairs$\downarrow$ & KV Access$\downarrow$ \\
  \midrule
  Dense
  & 86.2 & 5,213 & 0.0 & 1.000$\times$
  & Dense
  & 60.0 & 17,630 & 0.0 & 1.000$\times$ \\
  \addlinespace[1.5pt]
  \cdashline{1-5}[1pt/1.5pt]\cdashline{6-10}[1pt/1.5pt]
  \addlinespace[2pt]
  Quest
  & 67.6 & 13,025 & 0.0 & 1.230$\times$
  & Quest
  & 63.3 & 18,462 & 0.0 & 0.728$\times$ \\
  \addlinespace[4pt]
  ReSA-16
  & 82.6 & 8,225 & 513.5 & 0.888$\times$
  & ReSA-16
  & 70.0 & 17,812 & 1,112.5 & 0.764$\times$ \\
  \qfourName{0.5}
  & \textbf{85.6}
  & \qfourChange{6,624}{-19.5}
  & \qfourChange{408.6}{-20.4}
  & \qfourChange{0.642$\times$}{-27.7}
  & \qfourName{0.5}
  & \textbf{73.3}
  & \qfourChangeUp{18,296}{+2.7}
  & \qfourChange{1,084.9}{-2.5}
  & \qfourChangeUp{0.781$\times$}{+2.2} \\
  ReSA-32
  & 80.6 & 9,121 & 284.5 & 0.914$\times$
  & ReSA-32
  & 60.0 & 17,743 & 553.9 & 0.729$\times$ \\
  \qfourName{0.9}
  & 82.6
  & \qfourChange{7,663}{-16.0}
  & \qfourChange{241.2}{-15.2}
  & \qfourChange{0.729$\times$}{-20.2}
  & \qfourName{0.9}
  & 66.7
  & \qfourChange{17,646}{-0.5}
  & \qfourChange{514.4}{-7.1}
  & \qfourChange{0.717$\times$}{-1.6} \\
  \addlinespace[6pt]
  \midrule
  \multicolumn{5}{c}{\textbf{GPQA-Diamond} {\normalfont (198 questions)}}
  & \multicolumn{5}{c}{\textbf{LiveCodeBench v6} {\normalfont (175 questions)}} \\
  \cmidrule(lr){1-5}\cmidrule(lr){6-10}
  Method
  & Acc.$\uparrow$ & Tokens$\downarrow$ & Repairs$\downarrow$ & KV Access$\downarrow$
  & Method
  & Acc.$\uparrow$ & Tokens$\downarrow$ & Repairs$\downarrow$ & KV Access$\downarrow$ \\
  \midrule
  Dense
  & 55.1 & 8,904 & 0.0 & 1.000$\times$
  & Dense
  & 49.1 & 14,897 & 0.0 & 1.000$\times$ \\
  \addlinespace[1.5pt]
  \cdashline{1-5}[1pt/1.5pt]\cdashline{6-10}[1pt/1.5pt]
  \addlinespace[2pt]
  Quest
  & 50.5 & 14,076 & 0.0 & 1.123$\times$
  & Quest
  & 36.0 & 22,357 & 0.0 & 0.674$\times$ \\
  \addlinespace[4pt]
  ReSA-16
  & 55.6 & 10,119 & 631.9 & 0.836$\times$
  & ReSA-16
  & 48.6 & 17,648 & 1,102.4 & 0.603$\times$ \\
  \qfourName{0.7}
  & \textbf{56.1}
  & \qfourChange{9,599}{-5.1}
  & \qfourChange{587.9}{-7.0}
  & \qfourChange{0.776$\times$}{-7.2}
  & \qfourName{0.7}
  & \textbf{48.6}
  & \qfourChange{16,260}{-7.9}
  & \qfourChange{1,008.7}{-8.5}
  & \qfourChange{0.537$\times$}{-10.9} \\
  \addlinespace[4pt]
  ReSA-32
  & 52.5 & 10,615 & 331.2 & 0.848$\times$
  & ReSA-32
  & 48.0 & 18,259 & 569.9 & 0.584$\times$ \\
  \qfourName{1.1}
  & 55.1
  & \qfourChange{10,487}{-1.2}
  & \qfourChange{306.0}{-7.6}
  & \qfourChange{0.828$\times$}{-2.4}
  & \qfourName{1.1}
  & 47.4
  & \qfourChange{17,694}{-3.1}
  & \qfourChange{487.4}{-14.5}
  & \qfourChange{0.552$\times$}{-5.5} \\
  \bottomrule
  \end{tabular}%
  }
  \endgroup
  \vspace{-0.2cm}
\end{table*}

\begin{table*}[h!]
  \centering
  \caption{DeepSeek-R1-Distill-Llama-8B results. Notation follows Table~\ref{tab:qwen3_8b_main_results}.}
  \label{tab:llama_main_results}
  \vspace{3pt}
  \begingroup
  \small
  \setlength{\tabcolsep}{2pt}
  \renewcommand{\arraystretch}{1.12}

  \colorlet{llamaDelta}{green!60!black}
  \colorlet{llamaDeltaUp}{blue!75!black}

  \newcommand{\llamaChange}[2]{%
    #1\,\textcolor{llamaDelta}{\scriptsize(\ensuremath{#2}\%)}}
  \newcommand{\llamaChangeUp}[2]{%
    #1\,\textcolor{llamaDeltaUp}{\scriptsize(\ensuremath{#2}\%)}}
  \newcommand{\llamaName}[1]{%
    \textbf{\method}\,{\scriptsize
    \textcolor{black!60}{(#1)}}}

  \resizebox{\linewidth}{!}{%
  \begin{tabular}{@{}lllll@{\hspace{12pt}}lllll@{}}
  \toprule
  \multicolumn{5}{c}{\textbf{MATH-500} {\normalfont (500 questions)}}
  & \multicolumn{5}{c}{\textbf{AIME 2025} {\normalfont (30 questions)}} \\
  \cmidrule(lr){1-5}\cmidrule(lr){6-10}
  Method
  & Acc.$\uparrow$ & Tokens$\downarrow$ & Repairs$\downarrow$ & KV Access$\downarrow$
  & Method
  & Acc.$\uparrow$ & Tokens$\downarrow$ & Repairs$\downarrow$ & KV Access$\downarrow$ \\
  \midrule
  Dense
  & 83.0 & 4,356 & 0.0 & 1.000$\times$
  & Dense
  & 33.3 & 12,875 & 0.0 & 1.000$\times$ \\
  \addlinespace[1.5pt]
  \cdashline{1-5}[1pt/1.5pt]\cdashline{6-10}[1pt/1.5pt]
  \addlinespace[2pt]
  Quest
  & 75.9 & 8,155 & 0.0 & 0.836$\times$
  & Quest
  & 31.7 & 16,547 & 0.0 & 1.091$\times$ \\
  \addlinespace[4pt]
  ReSA-16
  & 82.5 & 4,853 & 302.7 & 0.517$\times$
  & ReSA-16
  & 38.3 & 13,352 & 834.0 & 0.911$\times$ \\
  \llamaName{0.7}
  & \textbf{83.7}
  & \llamaChange{4,673}{-3.7}
  & \llamaChange{266.6}{-11.9}
  & \llamaChange{0.482$\times$}{-6.8}
  & \llamaName{0.9}
  & \textbf{40.0}
  & \llamaChange{13,326}{-0.2}
  & \llamaChange{631.9}{-24.2}
  & \llamaChange{0.894$\times$}{-1.9} \\
  \addlinespace[4pt]
  ReSA-32
  & 81.2 & 5,734 & 178.7 & 0.596$\times$
  & ReSA-32
  & \textbf{40.0} & 16,041 & 500.7 & 1.097$\times$ \\
  \llamaName{1.3}
  & 81.7
  & \llamaChange{5,308}{-7.4}
  & \llamaChange{161.5}{-9.6}
  & \llamaChange{0.542$\times$}{-9.1}
  & \llamaName{1.2}
  & 35.0
  & \llamaChange{13,766}{-14.2}
  & \llamaChange{448.4}{-10.4}
  & \llamaChange{0.911$\times$}{-17.0} \\
  \midrule
  \multicolumn{5}{c}{\textbf{GPQA-Diamond} {\normalfont (198 questions)}}
  & \multicolumn{5}{c}{\textbf{LiveCodeBench v6} {\normalfont (175 questions)}} \\
  \cmidrule(lr){1-5}\cmidrule(lr){6-10}
  Method
  & Acc.$\uparrow$ & Tokens$\downarrow$ & Repairs$\downarrow$ & KV Access$\downarrow$
  & Method
  & Acc.$\uparrow$ & Tokens$\downarrow$ & Repairs$\downarrow$ & KV Access$\downarrow$ \\
  \midrule
  Dense
  & 40.7 & 9,440 & 0.0 & 1.000$\times$
  & Dense
  & 34.6 & 12,275 & 0.0 & 1.000$\times$ \\
  \addlinespace[1.5pt]
  \cdashline{1-5}[1pt/1.5pt]\cdashline{6-10}[1pt/1.5pt]
  \addlinespace[2pt]
  Quest
  & 40.7 & 11,908 & 0.0 & 0.726$\times$
  & Quest
  & 35.4 & 19,227 & 0.0 & 0.799$\times$ \\
  \addlinespace[4pt]
  ReSA-16
  & 44.2 & 8,368 & 522.5 & 0.514$\times$
  & ReSA-16
  & 34.3 & 11,487 & 717.4 & 0.485$\times$ \\
  \llamaName{0.8}
  & \textbf{44.4}
  & \llamaChange{8,276}{-1.1}
  & \llamaChange{515.8}{-1.3}
  & \llamaChange{0.505$\times$}{-1.8}
  & \llamaName{0.9}
  & \textbf{36.9}
  & \llamaChange{11,439}{-0.4}
  & \llamaChangeUp{730.3}{+1.8}
  & \llamaChange{0.478$\times$}{-1.4} \\
  \addlinespace[4pt]
  ReSA-32
  & 42.4 & 8,726 & 272.2 & 0.523$\times$
  & ReSA-32
  & 35.1 & 11,614 & 362.4 & 0.466$\times$ \\
  \llamaName{1.3}
  & 42.9
  & \llamaChange{8,532}{-2.2}
  & \llamaChange{265.0}{-2.6}
  & \llamaChange{0.505$\times$}{-3.4}
  & \llamaName{1.35}
  & 35.4
  & \llamaChangeUp{11,681}{+0.6}
  & \llamaChangeUp{370.5}{+2.2}
  & \llamaChange{0.466$\times$}{0.0} \\
  \bottomrule
  \end{tabular}%
  }
  \endgroup
  \vspace{-0.3cm}
\end{table*}

\paragraph{\method{} mitigates the degradation from sparse decoding.} Across three models and all four benchmarks (Tables~\ref{tab:qwen3_8b_main_results}, ~\ref{tab:qwen3_4b_main_results} and~\ref{tab:llama_main_results}), \method{} improves or matches Quest in accuracy while generating fewer tokens. The improvement is most pronounced largest on Qwen3-8B with MATH-500, where \method{} ($h=0.5$) increases accuracy from 67.0\% to 83.9\% while reducing the mean generation length from 13,274 to 6,932 tokens. Similar improvements hold across other tasks, showing that selectively repairing the accumulated KV errors substantially recovers the degradation caused by sparse decoding. The shorter generations can also offset the added refresh accesses: on Qwen3-4B/MATH-500, \method{} ($h=0.5$) has lower estimated KV-memory access than Quest (0.642$\times$ vs.\ 1.230$\times$ Dense).

\paragraph{Entropy-triggered refresh makes more effective use of dense correction.} At approximately matched refresh frequencies, \method{} achieves higher accuracy with shorter generations than ReSA. On Qwen3-4B/MATH-500, \method{} ($h=0.5$) further reduces estimated decoding KV-memory access by 27.7\% relative to ReSA-16 while improving accuracy by 3.0 points. \method{} ($h=0.9$) reduces access by 20.2\% while improving accuracy by 2.0 points. Thus, the gains are not explained by more frequent dense correction, but by \emph{when} it is applied. Targeting high-entropy steps makes better use of dense refresh. 

A \textbf{matched-random control} further supports our conclusion: at approximately matched refresh frequencies, \method{} improves accuracy by 3.2--3.8 points while generating 16.0--16.6\% fewer tokens than random refresh (Appendix~\ref{app:random-refresh}).

\subsection{The Pareto Advantage of Entropy-Triggered Repair}
\label{sec:accuracy_efficiency}

In this subsection, we show plots sweeping over the entropy threshold $h$ of \method{}, and compare it with sweeping the refresh frequency $r$ in periodic refresh to show that \method{} pareto dominates periodic refresh. 

\begin{figure}[h!]
    \centering
    \includegraphics[width=\linewidth]{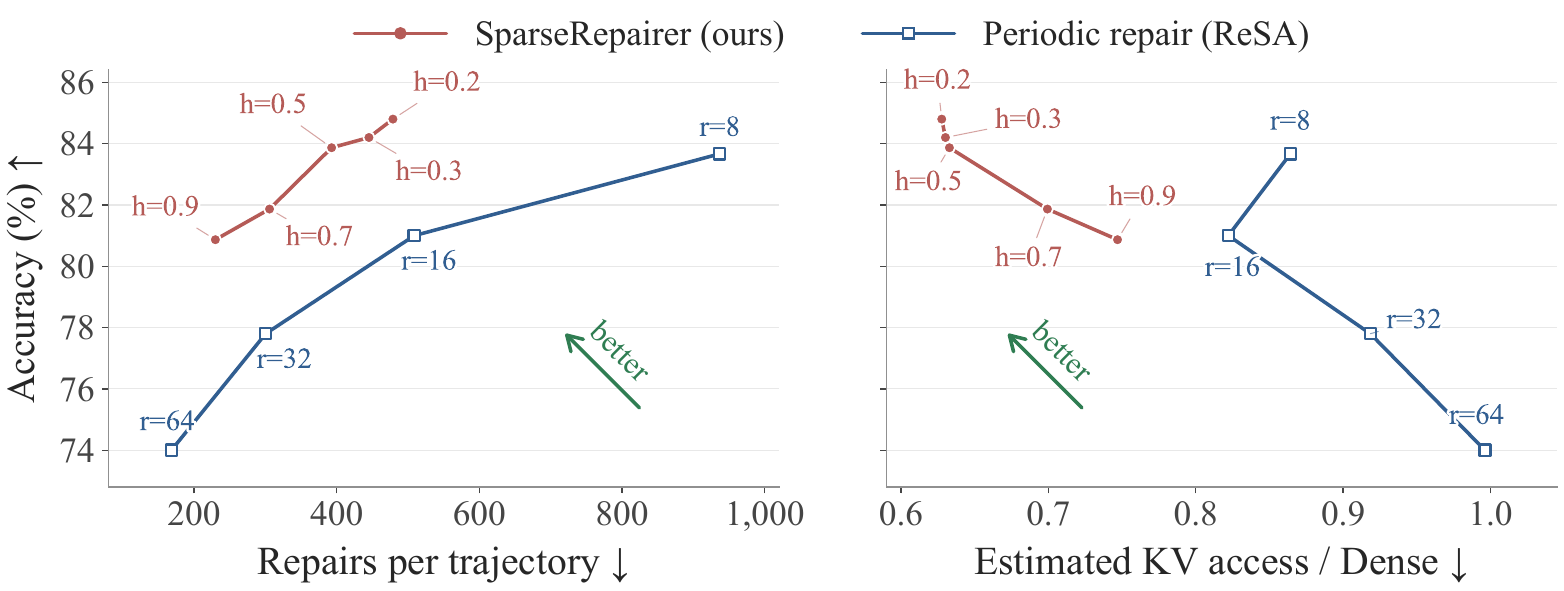}
    \includegraphics[width=\linewidth]{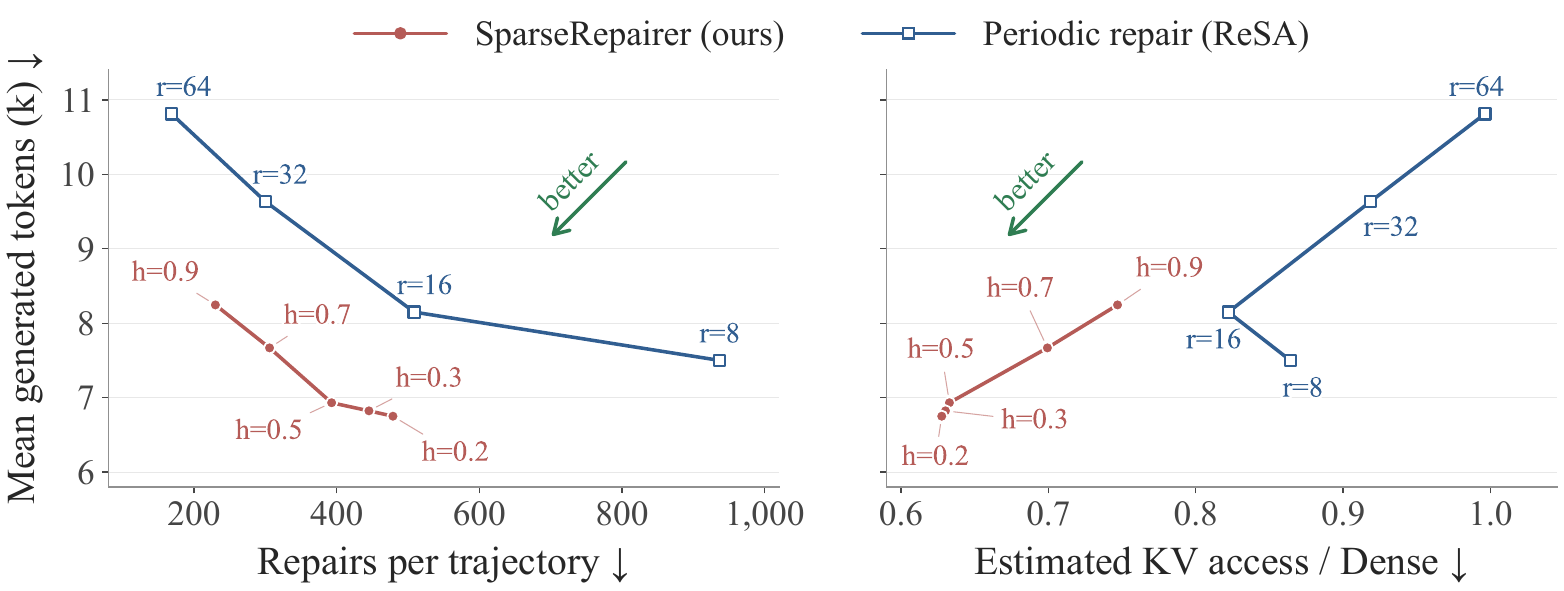}
    \caption{\textbf{\method{} Pareto-dominates periodic refresh with Quest.} It achieves higher accuracy (top) and shorter generations (bottom) at comparative refresh budgets or lower estimated KV access.}
    \label{fig:quest_tradeoff}
    \vspace{-0.3cm}
\end{figure}

\begin{figure}[h!]
    \centering
    \includegraphics[width=\linewidth]{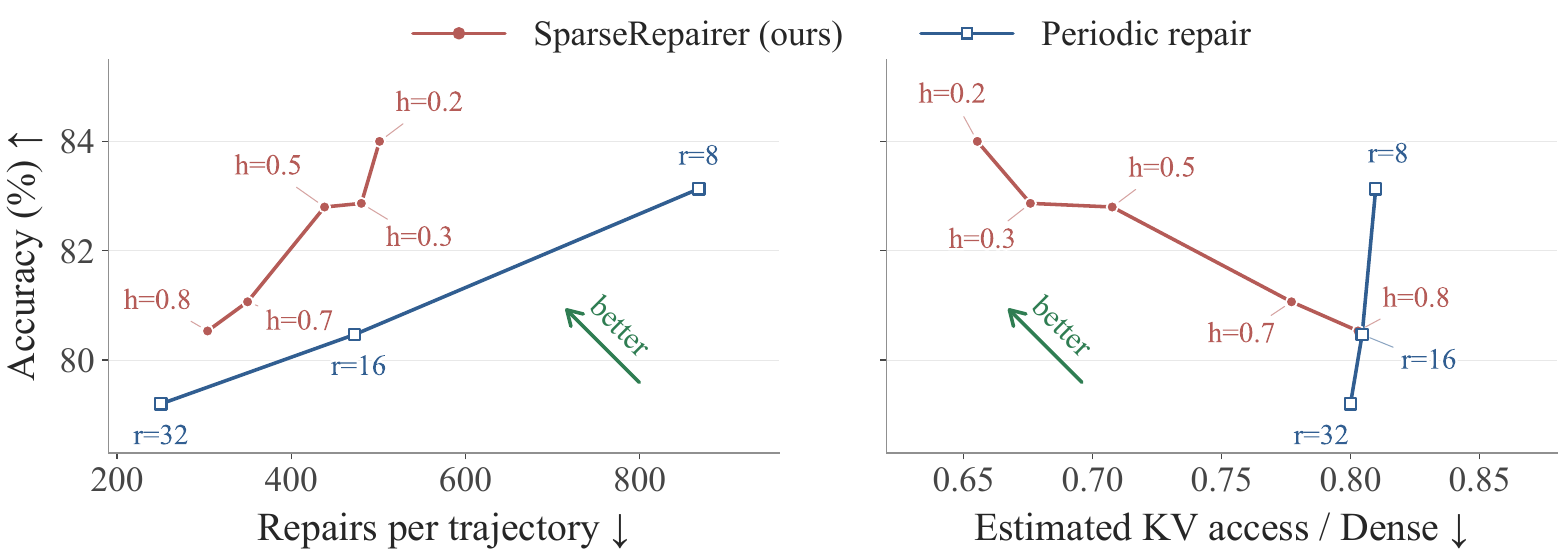}
    \includegraphics[width=\linewidth]{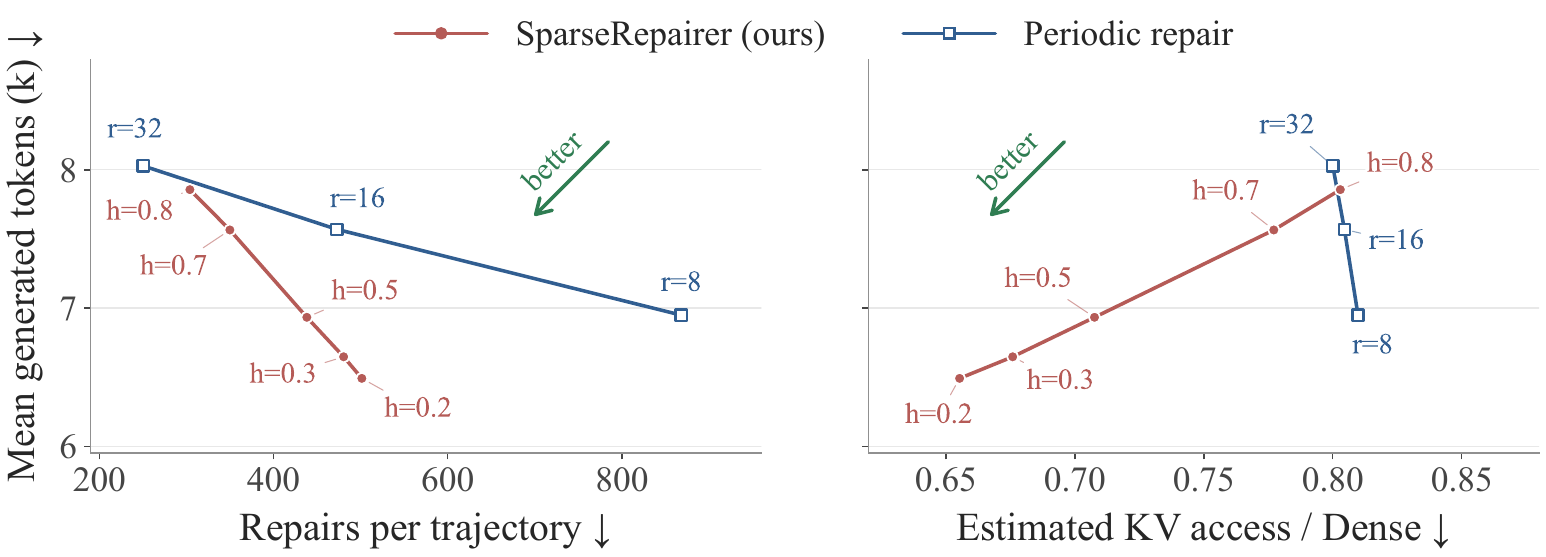}
    \caption{\textbf{\method{} retains its Pareto advantage with TidalDecode.} The gains in accuracy and generation length extend to a different sparse-attention mechanism.}
    \label{fig:tidal_tradeoff}
    \vspace{-0.2cm}
\end{figure}

We further compare \method{} with periodic refresh across different entropy thresholds and sparse-attention mechanisms. 
Specifically, we compare \method{} and periodic refresh under two different sparse attention methods: (1) Quest~\citep{tang2024quest} which selects KV pages based on estimated query-dependent scores and (2) TidalDecode~\citep{yang2025tidaldecode}, which reuses selected token positions across layers. We fix the sparse attention mechanism and only vary when to refresh to provide a clean comparison of refresh schedule.

Figures~\ref{fig:quest_tradeoff} and~\ref{fig:tidal_tradeoff} show that \method{} Pareto-dominates periodic refresh with both Quest and TidalDecode methods. \method{} achieves higher accuracy and shorter generations with fewer refresh calls and lower estimated KV access.

\subsection{Practical Inference Efficiency}
\label{sec:practical-efficiency}

\paragraph{Optimized implementation.}
To support efficient execution of \method{} in modern LLM inference systems, we build an optimized inference engine using paged KV caching, continuous batching, and CUDA Graphs. Both sparse decoding and dense KV refresh operate on a shared KV cache with 64-token pages to avoid additional KV reorganization when switching between them. For sparse decoding, we implement Quest page selection using custom Triton kernels and leverage FlashInfer paged attention to directly access selected KV pages. We further fuse KV-cache writes with page-level statistics updates to reduce maintenance overhead. For dense refresh, we recompute the unrepaired suffix with full causal attention in chunks of up to 64 tokens and update the KV cache in place. \method{} and periodic refresh share the same optimized engine, differing primarily in their refresh scheduling policies. 

\paragraph{Evaluation setup.}
We evaluate wall-clock inference efficiency using Qwen3-8B on Math87, a subset of 87 MATH-500 problems selected based on dense-decoding trajectories exceeding 8,000 tokens, independently of \method{}'s performance. We focus on these long-generation problems because both KV-cache access and dense refresh overhead become more significant as reasoning trajectories grow longer. All methods run on an NVIDIA RTX PRO 6000 GPU with a maximum concurrency of 14 and a 32,768-token generation limit. We report total wall-clock time averaged over two runs.

\begin{table}[h!]
    \centering
    \small
    \setlength{\tabcolsep}{5pt}
    \begin{tabular}{lrrrr}
        \toprule
        Method & Budget & Acc. (\%)$\uparrow$
        & Time (min)$\downarrow$ & Time Saved (\%)$\uparrow$ \\
        \midrule
        Dense & Full & 75.86 & 45.89 & -- \\
        \midrule
        Quest & 448 & 9.20 & 61.93 & $-35.0$ \\
        Quest & 1024 & 64.37 & 45.12 & $+1.7$ \\
        Quest & 2048 & 72.41 & 38.99 & $+15.0$ \\
        Quest & 3072 & 70.11 & 40.41 & $+11.9$ \\
        Quest & 4096 & 72.41 & 40.45 & $+11.9$ \\
        \midrule
        ReSA-16 & 448 & 48.28 & 56.47 & $-23.1$ \\
        ReSA-16 & 1024 & 72.41 & 38.20 & $+16.8$ \\
        ReSA-16 & 2048 & 74.71 & 38.09 & $+17.0$ \\
        \midrule
        \method{} ($h=0.5$) & 448 & 65.52 & 43.86 & $+4.4$ \\
        \method{} ($h=0.5$) & 1024 & 75.86 & 36.93 & $+19.5$ \\
        \textbf{\method{} ($h=0.5$)} & \textbf{2048}
        & \textbf{75.86} & \textbf{35.07} & \textbf{+23.6} \\
        \bottomrule
    \end{tabular}
    \caption{Wall-clock inference efficiency on Math87 with Qwen3-8B. Time is the total execution time for all 87 problems, averaged over two runs. Time Saved denotes the percentage reduction in wall-clock time relative to Dense; negative values indicate slower execution.}
    \label{tab:wallclock}
\end{table}

\paragraph{Wall-clock efficiency.}
As shown in Table~\ref{tab:wallclock}, increasing the sparse attention budget alone is insufficient: even with a 4,096-token budget, Quest achieves lower accuracy (72.41\% vs.\ 75.86\%) and higher wall-clock time (40.45 vs.\ 35.07 minutes) than \method{} with a 2,048-token budget. 

At the same 2,048-token budget, \method{} reduces wall-clock time by 7.9\% compared with periodic refresh while improving accuracy (75.86\% vs.\ 74.71\%). Moreover, \method{} matches dense decoding accuracy with 23.6\% lower wall-clock time.

\paragraph{Runtime overhead.}
We further profile the runtime overhead of \method{}. Entropy evaluation and trigger checking account for only 0.7\% of wall-clock time, while the estimated incremental forward cost of dense KV refresh accounts for 6.0\%. Detailed kernel-level profiling and overhead estimation are provided in Appendix~\ref{app:runtime-impl}.

\subsection{Impact of KV history refresh}
At high-entropy steps, \textbf{\method} removes inherited errors by repairing KV states with full attention, and removes local error by attending to all past tokens. We ablate the effect of repairing both types of errors: we compare \method{} with the following two variants:
\begin{itemize}[leftmargin=*]
    \item \method~with inherited errors: At high-entropy steps, we do not refresh the past KV but make the next token attend to all past tokens --- densely attending to a sparsely computed KV.
    \item \method~with local errors: At high-entropy steps, we do refresh the past KV but make the next token only attend to part of the dense KV --- sparsely attending to a densely computed KV.
\end{itemize}

The setup in this section differs from the results in Sec. \ref{sec:decomposition}. There, we use standard dense decoding at non-intervened steps, whereas in this section, we use sparse decoding (Quest).

Table~\ref{tab:kv_history_ablation} shows that both errors contribute to the performance degradation, but inherited error is more significant. 
\method{} with inherited errors reduces accuracy from 81.0\% to 74.0\%, whereas \method{} with local errors reduces it to 77.5\%.
The former also produces substantially longer generations (6,293 vs.\ 5,104 tokens) and a higher truncation rate (23.0\% vs.\ 15.0\%) than the latter.
These results are consistent with the inherited-error analysis in Sec.~\ref{sec:decomposition}. Repair also materially changes the resulting prediction: in a diagnostic analysis, 39--53\% of refresh events change the top-1 token, with mean distributional TV of 0.33--0.42 (Appendix~\ref{app:refresh-effect}).

\begin{table}[h!]
    \centering
    \small
    \setlength{\tabcolsep}{5pt}
    \caption{Impact of KV history refresh on the first 100 MATH-500
    problems with Qwen3-8B, using a maximum generation length of
    12,288 tokens. Results are averaged over two sampling seeds.
    $\Delta$ Tokens denotes the percentage change in generated
    tokens relative to full \method{}.}
    \label{tab:kv_history_ablation}
    \begin{tabular}{lcccc}
    \toprule
    Method & Acc. (\%) & Tokens & $\Delta$ Tokens (\%) & Trunc. (\%) \\
    \midrule
    Dense                          & 84.5 & 4,435 & -8.1  &  6.5 \\
    Quest \citep{tang2024quest}    & 66.0 & 6,839 & +41.7 & 31.5 \\
    \method{} with local errors     & 77.5 & 5,104 & +5.7  & 15.0 \\
    \method{} with inherited errors & 74.0 & 6,293 & +30.4 & 23.0 \\
    \textbf{\method{}} ($h=0.5$)    & 81.0 & 4,827 & -- & 12.5 \\
    \bottomrule
    \end{tabular}
\end{table}

\subsection{How Does \method{} Change Reasoning Trajectories?}
\label{sec:trajectory-analysis}

We analyze Qwen3-4B trajectories on MATH-500 to understand where \method{} changes the outcome. We compare \method{} ($h=0.5$) with Quest and ReSA-16 on the same 500 questions.
We call a baseline-incorrect/\method{}-correct pair a \emph{rescue}, and the reverse a \emph{regression}.

\paragraph{Most rescues involve truncated sparse trajectories.}
Relative to Quest, \method{} rescues 90 questions with no scored regressions, and 88 of these 90 rescued Quest trajectories reach the generation limit.
Relative to ReSA-16, \method{} rescues 20 questions and regresses on five, with 16 of the 20 rescues involving truncated ReSA-16 trajectories.
Thus, most observed rescues are associated with sparse trajectories that fail to terminate within the generation budget.

\paragraph{\method{} can avoid prolonged rechecking after reaching a solution.} For example, on MATH-500 problem 245 (Figure~\ref{fig:trajectory_math500} in Appendix~\ref{app:trajectory-analysis}), both Quest and \method{} derive the correct remainder, $-13x+3$, early in the trajectory. Quest subsequently repeats calculations and revisits previously verified steps until reaching the 32,768-token limit without completing its response. \method{} reaches the same result, performs additional checks, and terminates after 4,841 tokens. 

These trajectory-level observations do not imply that individual refresh events directly correct semantic reasoning errors. \method{} does not revise previously generated tokens, and some scored changes can also reflect answer formatting or extraction.

\section{Conclusion}

We study why sparse attention degrades long-form reasoning and find that its impact is concentrated at high-entropy decoding steps, where errors accumulated in the KV history account for most of the degradation.
We propose \method{}, which translates this diagnosis into an adaptive refresh policy that concentrates dense computation at these consequential steps. It uses sparse next-token entropy to selectively trigger dense KV refresh.
Across models and reasoning benchmarks, \method{} generally achieves higher accuracy and shorter generations than periodic refresh at comparable refresh frequencies.

\newpage
\subsection*{AI use statement}
We used generative AI tools to polish the writing of this paper, including grammar, wording, and readability. We also used generative AI to draft some portions of the appendix, all of which were reviewed and revised by the authors. The research ideas, methodology, experiments, and results are the authors' own. We take full responsibility for the content of this work.

\subsection*{Reproducibility statement}
We describe \method{} in Section~\ref{sec:sparse_repairer_method} and provide the experimental setup in Section~\ref{sec:experiments}, including the models (Qwen3-4B, Qwen3-8B, and DeepSeek-R1-Distill-Llama-8B), benchmarks, sparse attention settings, and all hyperparameters. For each model and benchmark, the entropy threshold $h$ is selected to match the refresh budget of ReSA, and the minimum refresh interval is fixed to $m=8$.

\section*{Acknowledgements}
This work was supported in part by ONR grant (N0001424-1-2089).
We acknowledge the use of computational resources on the Johns Hopkins Data Science and AI Institute (DSAI) cluster.

\newpage
\bibliography{iclr2027_conference}
\bibliographystyle{iclr2027_conference}

\newpage
\appendix
\section{Appendix}

\subsection{Analytical Memory-Access Estimate}
\label{app:memory-access}

We estimate decoding KV-memory access using an analytical model adapted from ReSA~\citep{sun2026rectified}.
Our goal is to compare Quest, periodic refresh, and \method{} under a common sparse-attention implementation rather than reproduce implementation-specific kernel behavior.
All sparse methods therefore use the same Quest-style backend with per-query-head page selection.
The analysis below provides a common accounting framework for the resulting decoding trajectories and refresh schedules.

\paragraph{Cost model.}
We express memory access in KV-token equivalents per KV head, omitting common factors from layers, head dimension, and element size.
Let $g=H_q/H_{\mathrm{KV}}$ denote the number of query heads sharing each KV head; $g=4$ for all models evaluated in this work.
For decoding step $t$, let $L_t$ denote the current KV-cache length, $b$ the page size, and $B=kb$ the sparse-attention budget.

Our sparse backend performs page selection independently for each query head, whereas dense attention and dense rectification can reuse the KV states within a GQA group.
For an ordinary decoding forward of a sparse method, we therefore estimate
\begin{equation}
C_t^{\mathrm{sparse}} =
\begin{cases}
L_t, & L_t \le B,\\
g\left(\frac{L_t}{b}+A_t\right), & L_t > B,
\end{cases}
\label{eq:sparse-access}
\end{equation}
where the first case accounts for the backend's dense fallback on short contexts.
For $L_t>B$, the backend selects $k$ pages per query head and always retains the final page.
The number of valid selected KV tokens per query head is therefore reconstructed as
\begin{equation}
A_t=(k-1)b+1+((L_t-1)\bmod b).
\label{eq:selected-kv}
\end{equation}
This quantity reconstructs selection cardinality from the page-selection rule rather than logged page indices.
In particular, it does not measure overlap or reuse of selected pages across query heads.

Following ReSA's batched-rectification abstraction, we count each dense refresh as one additional full-history KV access.
For a trajectory with ordinary decoding forwards $\mathcal{D}$ and refresh positions $\mathcal{R}$,
\begin{equation}
C_{\mathrm{mem}}
=
\sum_{t\in\mathcal{D}} C_t^{\mathrm{sparse}}
+
\sum_{r\in\mathcal{R}} L_r.
\label{eq:memory-access}
\end{equation}
In our recorded runs, $|\mathcal{D}|=G-1$ for a trajectory with $G$ generated tokens because the first token is sampled from the prefill output.
For \method{}, the sparse forward used to compute the triggering entropy is included even when it is immediately followed by a refresh.

For dense decoding, we use
\begin{equation}
C_{\mathrm{mem}}^{\mathrm{Dense}}
=
\sum_{t\in\mathcal{D}} L_t.
\label{eq:dense-access}
\end{equation}
We exclude the common dense prefill from all methods while including the prompt KV states in $L_t$ during subsequent decoding.
We aggregate the estimated access over all trajectories and normalize by the corresponding aggregate Dense cost.

This is an analytical access estimate rather than a measurement of physical memory traffic or latency.

\subsection{Matched-Random Repair Control}
\label{app:random-refresh}

To isolate the effect of \emph{where} refresh is applied, we compare \method{} against a random-trigger refresh policy on Qwen3-4B and MATH-500.
For each \method{} threshold, the random policy is calibrated to the same trigger frequency, while using the same minimum refresh interval and dense-replay implementation.
Both policies reuse the logits produced by dense replay for next-token sampling.

Table~\ref{tab:random-refresh} reports results over all 500 problems with one sampling seed.
The realized refresh frequencies are nearly identical: 6.158\% versus 6.170\% for the $h=0.5$ pair, and 3.145\% versus 3.149\% for the $h=0.9$ pair.
Despite this, \method{} consistently achieves higher accuracy and shorter generations.
At $h=0.5$, \method{} improves accuracy by 3.2 percentage points and reduces generation length by 16.0\%; at $h=0.9$, it improves accuracy by 3.8 points and reduces generation length by 16.6\%.
Because the resulting trajectories are shorter, \method{} also requires fewer total refreshes.
These results provide direct evidence that the gains come from \emph{where} dense refresh is applied rather than from repairing more frequently.

\begin{table}[t]
    \centering
    \small
    \setlength{\tabcolsep}{5pt}
    \caption{
    \textbf{Entropy-triggered refresh outperforms random refresh at matched refresh frequencies.}
    Results on Qwen3-4B and MATH-500 over all 500 problems.
    Repair frequency denotes refresh calls per generated token.
    }
    \label{tab:random-refresh}
    \begin{tabular}{lrrrr}
        \toprule
        Method & Acc. (\%)$\uparrow$ & Tokens$\downarrow$ & Repairs$\downarrow$ & Freq. (\%) \\
        \midrule
        Random, matched to $h=0.5$ & 82.4 & 7,884 & 485.4 & 6.158 \\
        \method{} ($h=0.5$)             & 85.6 & 6,624 & 408.6 & 6.170 \\
        \midrule
        Random, matched to $h=0.9$ & 78.8 & 9,190 & 289.0 & 3.145 \\
        \method{} ($h=0.9$)             & 82.6 & 7,663 & 241.2 & 3.149 \\
        \bottomrule
    \end{tabular}
\end{table}

\subsection{What Changes When \method{} Repairs?}
\label{app:refresh-effect}
Section~\ref{sec:mechanism} motivates refresh at high-entropy positions, but it does not show how refresh changes the next-token distribution.
Moreover, Section~\ref{sec:mechanism} analyzes fixed dense trajectories, whereas \method{} refreshes its own trajectories, in which earlier refreshes have already restored part of the cache.
In this section, we therefore analyze how \method{} refreshes change the next-token distribution along \method{} trajectories.

\paragraph{Setup and measurement.} We analyze \method{} trajectories generated by Qwen3-8B on GPQA-Diamond with thresholds $h = 0.5$ and $h = 0.9$, using the first 118 questions. We use a Quest budget of 448 tokens and a 12{,}288-token generation limit, and \method{} enforces a minimum interval of 8 tokens between consecutive refreshes. Sampling uses temperature 0.6 and top-$p$ 0.95. This setting differs from the main GPQA-Diamond experiment (Section~\ref{sec:experimental_setup}), which uses a 1{,}024-token budget and a 32{,}768-token limit on all 198 questions.

At a refresh-triggering position $t$, let $p_t^-$ be the sparse next-token distribution before refresh, and let $p_t^+$ be the dense distribution obtained after recomputing the KV states of the tokens generated since the previous refresh. Both are full-vocabulary distributions at temperature 0.6 before top-$p$ truncation. We compute
\begin{equation}
 H_t^{\pm}=-\sum_{v\in\mathcal V}p_t^{\pm}(v)\log p_t^{\pm}(v),
 \qquad \Delta H_t=H_t^- - H_t^+,
 \qquad \mathrm{TV}_t = \tfrac{1}{2}\sum_{v\in\mathcal V}\left|p_t^-(v)-p_t^+(v)\right|,
\end{equation}
where a positive $\Delta H_t$ means that refresh lowers the next-token entropy, and $\mathrm{TV}_t$ measures how much refresh changes the distribution.

\paragraph{Repair lowers entropy on average but does not remove it.}
Table~\ref{tab:refresh-distributions} summarizes the effect of refresh. Entropy decreases in 66.5\% and 75.5\% of refreshes for $h=0.5$ and $h=0.9$, respectively (mean $\Delta H_t = 0.1710$ and $0.2999$), and 32.7\% and 51.0\% of refreshes end below the threshold, i.e., the sparse entropy that triggered them was at least partly induced by sparse error.

\paragraph{Repair often changes the sampled continuation.}
The top-1 token changes in 39.3\% and 53.0\% of refreshes, with mean TV of 0.3266 and 0.4170. Because \method{} samples from $p_t^+$, these changes directly alter the trajectory.

\paragraph{Repaired positions remain uncertain.}
Across all \method{} decoding steps, mean entropy is 0.2310 ($h=0.5$) and 0.2285 ($h=0.9$), whereas post-refresh entropy at triggered positions is 0.6686 and 0.8587; 84.8\% and 90.7\% of repaired positions remain above the all-step mean. Thus \method{} spends dense computation on intrinsically uncertain steps rather than only removing sparsity-induced uncertainty, consistent with Section~\ref{sec:mechanism}. Two caveats apply: because refreshes are selected for high pre-refresh entropy, part of the observed decrease reflects regression to the mean; and without a fully dense reference at every prefix, these results do not show that sparse entropy exactly recovers dense-model uncertainty.

\begin{table}[t]
\centering
\small
\caption{\textbf{Repair reduces entropy on average and frequently changes the leading candidate.} Effect of \method{} refreshes for Qwen3-8B on 118 GPQA-Diamond questions. \textit{Repairs}: total number of refreshes. \textit{Mean $H_t^-\!\to\!H_t^+$}: mean next-token entropy before and after refresh. \textit{All-step $H$}: mean next-token entropy over all \method{} decoding steps. \textit{Decreased}: percentage of refreshes after which entropy decreased. \textit{Below $h$}: percentage of refreshes whose post-refresh entropy falls below the threshold $h$. \textit{Argmax changed}: percentage of refreshes that change the top-1 token. \textit{Mean TV}: mean total variation distance between the full-vocabulary distributions before and after refresh.}
\label{tab:refresh-distributions}
\setlength{\tabcolsep}{4pt}
\resizebox{\linewidth}{!}{%
\begin{tabular}{lrrrrrrr}
\toprule
$h$ & Repairs & Mean $H_t^-\!\to\!H_t^+$ & All-step $H$ & Decreased & Below $h$ & Argmax changed & Mean TV \\
\midrule
0.5 & 88,768 & $0.8396\to0.6686$ & 0.2310 & 66.5\% & 32.7\% & 39.3\% & 0.3266 \\
0.9 & 52,229 & $1.1586\to0.8587$ & 0.2285 & 75.5\% & 51.0\% & 53.0\% & 0.4170 \\
\bottomrule
\end{tabular}%
}
\end{table}

\subsection{How does \method{} choose $h$?}
\label{app:choosing_h}
For configurations using frequency-based calibration, we collect frozen Quest trajectories on a separate unlabeled calibration set.
For each candidate threshold $h$, we replay the \method{} gate ($H_t^S \geq h$) with the same minimum refresh interval $m=8$ used at inference time, and count the resulting refresh events.
We select $h$ to approximately match the refresh frequency of ReSA-16 or ReSA-32 on trajectories of the same lengths.
The selected threshold is then fixed for evaluation.
This procedure uses neither correctness labels nor \method{} outputs.

\subsection{Full Results of the Intervention Study}
\label{app:intervention}

Table~\ref{tab:math500-entropy-sweep} reports the full results of the intervention study in Sec.~\ref{sec:entropy_sensitivity}, including accuracy, generation length, and truncation rate at each entropy threshold.
Across thresholds, high-entropy interventions consistently cause substantially greater degradation than approximately rate-matched random interventions.

\begin{table*}[t]
    \centering
    \small
    \setlength{\tabcolsep}{7pt}
    \renewcommand{\arraystretch}{1.15}
    \caption{\textbf{Sparse attention is substantially more harmful at high-entropy steps than at randomly selected steps with similar intervention rates.} Results on MATH-500 with Qwen3-8B and Quest (token budget 448). Sparse attention is applied only at intervened steps, while all other steps use full attention. High-entropy interventions are selected by $H_t \geq \tau$; random interventions are approximately matched to their intervention rates. Results are averaged over three seeds with a 12,288-token generation limit.}
    \label{tab:math500-entropy-sweep}
    \resizebox{\textwidth}{!}{%
    \begin{tabular}{lrrrrrrrr}
    \toprule
    & \multicolumn{4}{c}{High entropy}
    & \multicolumn{4}{c}{Approximately rate-matched random} \\
    \cmidrule(lr){2-5} \cmidrule(lr){6-9}
    $\tau$
    & Rate (\%)
    & Acc.\ (\%)$\uparrow$
    & Tokens$\downarrow$
    & Trunc.\ (\%)$\downarrow$
    & Rate (\%)
    & Acc.\ (\%)$\uparrow$
    & Tokens$\downarrow$
    & Trunc.\ (\%)$\downarrow$ \\
    \midrule
    \textbf{Dense}
        & 0.00 & \hmc{100}82.47 & \hmc{100}4,953 & 8.13
        & \multicolumn{4}{c}{--} \\
    \midrule
    \hmc{20}0.2 & 23.75 & \hmc{20}64.70 & \hmc{20}6,909 & 29.50
        & 23.74 & 79.40 & 5,465 & 12.50 \\
    \hmc{35}0.5 & 14.67 & \hmc{35}71.73 & \hmc{35}6,360 & 22.47
        & 14.52 & 81.07 & 5,261 & 10.33 \\
    \hmc{50}0.7 & 7.71 & \hmc{50}76.00 & \hmc{50}5,924 & 17.10
        & 7.71 & 82.80 & 5,130 & 9.30 \\
    \hmc{65}0.9 & 4.00 & \hmc{65}80.27 & \hmc{65}5,478 & 12.13
        & 3.99 & 82.73 & 5,015 & 8.87 \\
    \bottomrule
    \end{tabular}%
    }
\end{table*}

\subsection{Which Tokens Trigger Repair?}
\label{app:trigger-tokens}
We also examine which tokens precede refresh, i.e., the last generated token before each refresh-triggering position. Frequent tokens appear often at refresh positions simply because they are frequent, so raw counts alone are uninformative. We therefore also report a relative frequency. For a token $v$, let $f_r(v)$ be the fraction of refresh-triggering positions preceded by $v$, and let $f_n(v)$ be the corresponding fraction among non-refresh positions. We define
\begin{equation}
F(v) = \frac{f_r(v)}{f_n(v)},
\end{equation}
so $F(v) > 1$ means that $v$ precedes refreshes more often than its base rate. Table~\ref{tab:refresh-trigger-tokens} lists the most frequent refresh-preceding tokens, and Table~\ref{tab:refresh-token-f} lists the tokens with the highest $F(v)$. By count, refreshes are most often preceded by punctuation, paragraph breaks, and function words (``.'', ``,'', ``the'', ``is''). By relative frequency, the most over-represented tokens signal a shift in reasoning: contrast (``But'', ``but'', ``Or''), verification (``check'', ``typo'', ``calculation'', ``seems''), and conclusion (``So''). This is consistent with prior findings that high-entropy tokens act as forks in reasoning~\citep{wang2025beyond}. In contrast, digits such as ``2'' are under-represented ($F = 0.42$ and $0.24$), suggesting that routine computation rarely triggers refresh.

\begin{table}[t]
\centering
\small
\setlength{\tabcolsep}{4pt}
\caption{\textbf{Most frequent refresh-preceding tokens.} Tokens are sorted by their combined refresh count over both thresholds. $F$ is the relative frequency defined in Appendix~\ref{app:trigger-tokens}. Repairs are most often preceded by punctuation (e.g., ``.'', ``,'', ``?''), conjunctions (e.g., ``So'', ``but''), and function words (e.g., ``the'', ``is'', ``to'').}
\label{tab:refresh-trigger-tokens}
\begin{tabular}{lrrrr}
\toprule
Preceding token & $h=0.5$ count & $F$ & $h=0.9$ count & $F$ \\
\midrule
\verb|"."| & 9,835 & $3.28\times$ & 7,169 & $3.43\times$ \\
\verb|","| & 8,863 & $2.89\times$ & 7,070 & $3.19\times$ \\
\verb|" the"| & 4,991 & $2.19\times$ & 3,831 & $2.09\times$ \\
\verb|" is"| & 2,774 & $1.96\times$ & 1,677 & $1.63\times$ \\
\verb|" \n\n"| & 2,216 & $6.99\times$ & 1,896 & $5.59\times$ \\
\verb|" a"| & 1,530 & $1.49\times$ & 1,064 & $1.50\times$ \\
\verb|"?"| & 1,279 & $5.09\times$ & 924 & $3.59\times$ \\
\verb|" So"| & 901 & $5.78\times$ & 846 & $4.43\times$ \\
\verb|" to"| & 1,051 & $1.46\times$ & 594 & $1.26\times$ \\
\verb|"2"| & 1,294 & $0.42\times$ & 343 & $0.24\times$ \\
\verb|" But"| & 720 & $7.81\times$ & 914 & $7.32\times$ \\
\verb|" but"| & 733 & $5.51\times$ & 859 & $5.51\times$ \\
\bottomrule
\end{tabular}
\end{table}

\begin{table}[t]
\centering
\small
\setlength{\tabcolsep}{4pt}
\caption{\textbf{Tokens with the highest relative frequency $F$.} Tokens appearing fewer than 10 times in either the refresh or the non-refresh set are excluded. The most over-represented tokens mark shifts in reasoning, such as contrast (``But''), verification (``check'', ``typo''), and conclusion (``So''). ``rotein'' is a subword of ``protein''.}
\label{tab:refresh-token-f}
\begin{tabular}{rlrrr}
\toprule
Rank & Preceding token & Repair count & Non-refresh count & $F$ \\
\midrule
\multicolumn{5}{c}{$h=0.5$} \\
\midrule
1  & \verb|"But"|         & 348   & 173   & $9.72\times$ \\
2  & \verb|" But"|        & 720   & 446   & $7.81\times$ \\
3  & \verb|" check"|      & 193   & 126   & $7.40\times$ \\
4  & \verb|" \n\n"|       & 2,216 & 1,534 & $6.99\times$ \\
5  & \verb|" count"|      & 14    & 10    & $6.68\times$ \\
6  & \verb|" calculated"| & 21    & 16    & $6.30\times$ \\
7  & \verb|" Because"|    & 23    & 18    & $6.15\times$ \\
8  & \verb|"\"."|        & 39    & 31    & $6.07\times$ \\
9  & \verb|" So"|         & 901   & 754   & $5.78\times$ \\
10 & \verb|" Or"|         & 74    & 64    & $5.59\times$ \\
11 & \verb|" should"|     & 106   & 93    & $5.51\times$ \\
12 & \verb|" but"|        & 733   & 644   & $5.51\times$ \\
\midrule
\multicolumn{5}{c}{$h=0.9$} \\
\midrule
1  & \verb|"But"|         & 571   & 794   & $10.31\times$ \\
2  & \verb|" But"|        & 914   & 1,791 & $7.32\times$ \\
3  & \verb|"rotein"|      & 11    & 25    & $6.46\times$ \\
4  & \verb|" see"|        & 23    & 52    & $6.41\times$ \\
5  & \verb|" should"|     & 115   & 267   & $6.19\times$ \\
6  & \verb|"**\n\n"|      & 16    & 39    & $5.99\times$ \\
7  & \verb|" \n\n"|       & 1,896 & 4,864 & $5.59\times$ \\
8  & \verb|" but"|        & 859   & 2,233 & $5.51\times$ \\
9  & \verb|" typo"|       & 14    & 42    & $4.89\times$ \\
10 & \verb|" seems"|      & 28    & 87    & $4.67\times$ \\
11 & \verb|")."|          & 354   & 1,116 & $4.55\times$ \\
12 & \verb|" calculation"| & 21  & 68    & $4.50\times$ \\
\bottomrule
\end{tabular}
\end{table}



\subsection{Additional Trajectory Analysis}
\label{app:trajectory-analysis}

\paragraph{Case study: repeated verification after obtaining the answer.}
We examine MATH-500 problem 245(\texttt{math500\_245}) using Qwen3-4B. The task asks for the remainder of a polynomial $f(x)$ upon division by $x^2-1$. Writing the remainder as $ax+b$, evaluating at $x=1$ and $x=-1$ gives $a+b=-10$ and $-a+b=16$. Thus, $a=-13$ and $b=3$, which fully determines the correct remainder, $-13x+3$.

\begin{figure}[t]
    \centering
    \includegraphics[width=\linewidth, 
    ]{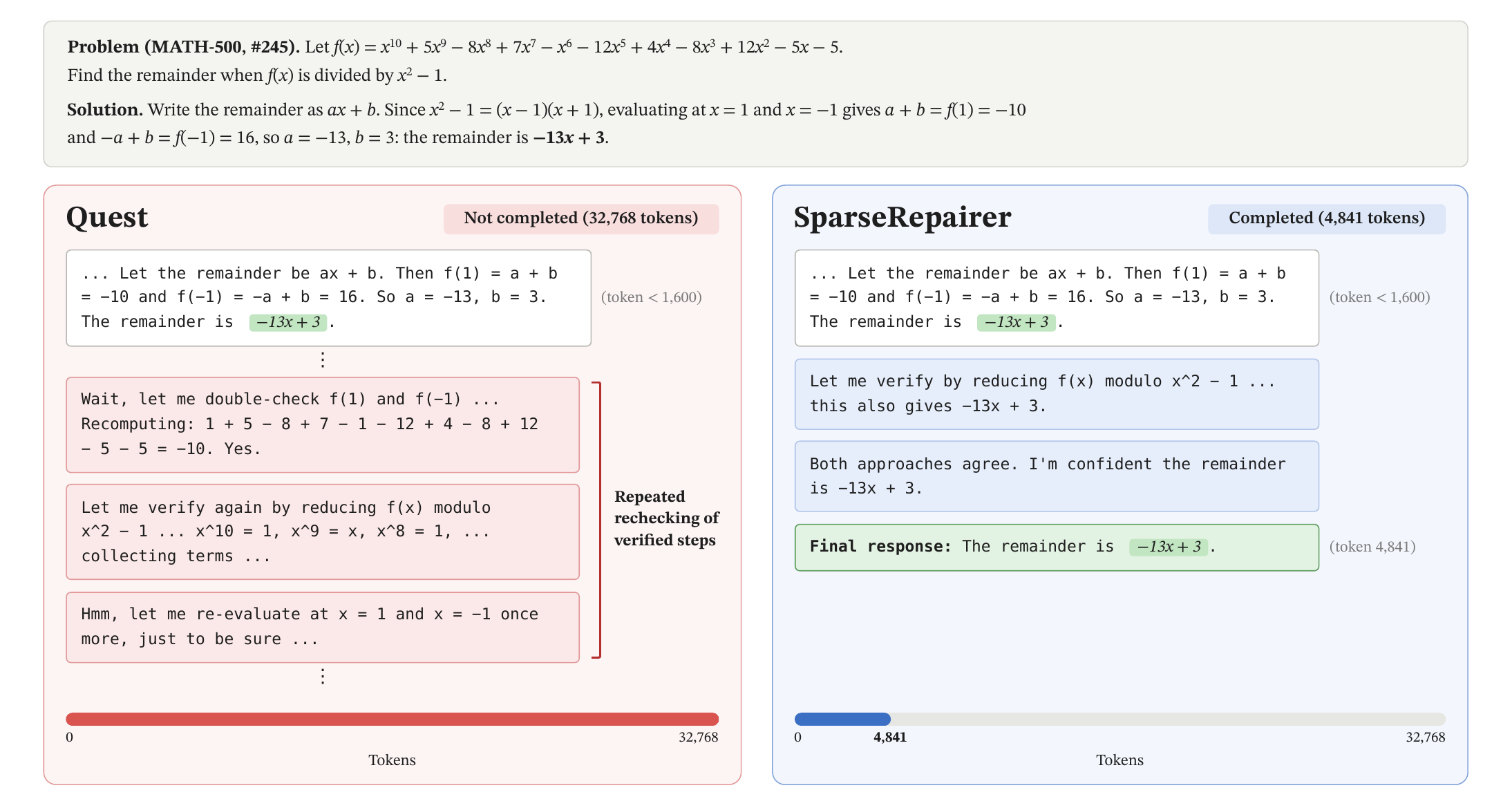}
    \caption{Case study on MATH-500 problem 245 with Qwen3-4B. Both Quest (left) and \method{} (right) obtain the correct remainder $-13x+3$ within the first 1,600 generated tokens. Quest keeps rechecking previously verified steps until it reaches the 32,768-token limit without completing its response; \method{} verifies the result once more and completes its response at 4,841 tokens. Excerpts are abridged; bars show generated tokens relative to the limit.}
    \label{fig:trajectory_math500}
\end{figure}

The excerpts in Figure~\ref{fig:trajectory_math500} distinguish obtaining this result during thinking from completing the final response. Both methods obtain the correct remainder within the first 1,600 generated tokens. Quest subsequently revisits the same evaluations and polynomial reductions, and is still checking when it reaches the 32,768-token limit. \method{} also verifies its result, but proceeds to a correct final response and terminates after 4,841 tokens.  These excerpts illustrate the behavior of the two complete methods, rather than the effect of an isolated refresh.

\subsection{Truncation Rate}

\autoref{tab:truncation_rates} shows the truncation rates for all models and benchmarks in our study. We found that \method{} substantially reduces truncation relative to Quest and often improves over periodic refresh. This indicates that \method{} is not only better at lowering average length of generation, it also alleviates the long-tailed generation induced by sparse attention.

\label{app:truncation}
\begin{table*}[t]
  \centering
  \caption{Truncation rate (\%) of all methods across models and benchmarks, complementing Tables~\ref{tab:qwen3_8b_main_results}--\ref{tab:llama_main_results}. Parenthesized values beside \method{} denote the entropy threshold $h$; the first \method{} column pairs with ReSA-16 and the second with ReSA-32.}
  \label{tab:truncation_rates}
  \begingroup
  \small
  \setlength{\tabcolsep}{6pt}
  \renewcommand{\arraystretch}{1.12}
  \newcommand{\truncH}[2]{#1\,{\scriptsize\textcolor{black!60}{(#2)}}}
  \begin{tabular}{@{}lllllll@{}}
  \toprule
  Benchmark & Dense & Quest & ReSA-16 & \textbf{\method{}} & ReSA-32 & \textbf{\method{}} \\
  \midrule
  \multicolumn{7}{@{}l}{\textit{Qwen3-8B}} \\
  MATH-500 & 8.1 & 36.8 & 20.7 & \truncH{15.1}{0.5} & 23.9 & \truncH{20.5}{0.9} \\
  AIME 2025 & 10.0 & 28.9 & 16.7 & \truncH{11.1}{0.5} & 16.7 & \truncH{17.8}{0.9} \\
  GPQA-Diamond & 0.0 & 14.9 & 1.5 & \truncH{0.3}{0.5} & 1.0 & \truncH{1.3}{1.1} \\
  LiveCodeBench v6 & 2.6 & 53.4 & 14.6 & \truncH{12.6}{0.9} & 23.4 & \truncH{16.6}{1.05} \\
  \addlinespace[4pt]
  \multicolumn{7}{@{}l}{\textit{Qwen3-4B}} \\
  MATH-500 & 0.0 & 25.0 & 8.2 & \truncH{3.4}{0.5} & 11.0 & \truncH{7.0}{0.9} \\
  AIME 2025 & 10.0 & 23.3 & 20.0 & \truncH{10.0}{0.5} & 20.0 & \truncH{13.3}{0.9} \\
  GPQA-Diamond & 1.0 & 11.6 & 0.5 & \truncH{0.5}{0.7} & 2.5 & \truncH{2.0}{1.1} \\
  LiveCodeBench v6 & 2.9 & 56.0 & 17.1 & \truncH{8.0}{0.7} & 27.4 & \truncH{21.1}{1.1} \\
  \addlinespace[4pt]
  \multicolumn{7}{@{}l}{\textit{DeepSeek-R1-Distill-Llama-8B}} \\
  MATH-500 & 0.6 & 12.7 & 1.9 & \truncH{1.1}{0.7} & 3.4 & \truncH{2.0}{1.3} \\
  AIME 2025 & 5.0 & 11.7 & 5.0 & \truncH{3.3}{0.9} & 8.3 & \truncH{3.3}{1.2} \\
  GPQA-Diamond & 5.1 & 8.1 & 0.0 & \truncH{0.0}{0.8} & 0.8 & \truncH{0.5}{1.3} \\
  LiveCodeBench v6 & 6.9 & 27.4 & 0.3 & \truncH{0.3}{0.9} & 0.9 & \truncH{0.9}{1.35} \\
  \bottomrule
  \end{tabular}
  \endgroup
\end{table*}

\subsection{Additional Intervention Study on GPQA-Diamond}
\label{app:gpqa-intervention}

To test whether the high-entropy sensitivity observed on MATH-500 extends beyond mathematics, we repeat the intervention study of Sec.~3.2 on GPQA-Diamond with Qwen3-8B.
We compare sparse-history/sparse-current interventions at positions with $H_t \geq 0.5$ against an approximately rate-matched random control, while otherwise decoding with full attention.
We use a Quest budget of 1,024 tokens, a 32,768-token generation limit, and one sampling seed.
Both policies intervene at roughly 19\% of generated positions.

As shown in Table~\ref{tab:gpqa_intervention}, high-entropy interventions substantially lengthen generation relative to both dense decoding and the random control, whereas random interventions have little effect.
The corresponding accuracy differences are small.
Together with the MATH-500 results, this provides additional evidence that sparse states at high-entropy decisions have disproportionately large effects on reasoning trajectories.

\begin{table}[t]
  \centering
  \caption{\textbf{High-entropy interventions lengthen generations on GPQA-Diamond far more than rate-matched random interventions.} Results on GPQA-Diamond with Qwen3-8B and Quest (token budget 1,024) with 32,768-token generation limit. Sparse attention is applied only at intervened steps, while all other steps use full attention. High-entropy interventions are selected by $H_t \geq 0.5$; random interventions are approximately matched to the same intervention rate. Rate is the fraction of intervened steps.}
  \label{tab:gpqa_intervention}
  \setlength{\tabcolsep}{6pt}
  \renewcommand{\arraystretch}{1.12}
  \resizebox{\ifdim\width>\linewidth\linewidth\else\width\fi}{!}{%
  \begin{tabular}{@{}lrrrr@{}}
  \toprule
  Policy & Acc. (\%) & Tokens & Trunc. (\%) & Rate (\%) \\
  \midrule
  Dense & 61.62 & 9,825 & 0.00 & -- \\
  High entropy & 60.10 & 12,508 & 1.52 & 19.74 \\
  Matched random & 62.12 & 10,226 & 0.51 & 19.25 \\
  \bottomrule
  \end{tabular}}
\end{table}

\section{Optimized Implementation and Runtime Profiling}
\label{app:runtime-impl}

\paragraph{GPU implementation.}
Our optimized inference engine supports both sparse decoding and dense KV refresh through a shared paged KV cache with 64-token pages. During sparse decoding, custom Triton kernels fuse KV-cache writes with page-level key-statistics updates, avoiding separate kernels for maintaining Quest's page metadata. Quest page selection is also performed on the GPU, and the selected pages are accessed directly through FlashInfer paged attention without gathering KV states into contiguous buffers.

When \method{} triggers a refresh, the engine recomputes the unrepaired suffix using full causal attention in chunks of up to 64 tokens. The recomputed KV states replace the corresponding entries in the existing paged cache, allowing subsequent sparse decoding to reuse the same cache layout without additional KV reorganization. The engine supports continuous batching and CUDA Graph execution, enabling sparse decoding and refresh within the same inference runtime. \method{} and periodic refresh use the same optimized engine, differing primarily in their refresh scheduling policies.

\paragraph{Profiling setup.}
We profile \method{} with $h=0.5$ and a 2,048-token attention budget on Qwen3-8B and Math87, using an NVIDIA RTX PRO 6000 GPU. The experiment uses a maximum concurrency of 14 and a generation limit of 32,768 tokens. We collect the complete execution trace, containing 95,174 scheduler steps and approximately 2,012 seconds of GPU kernel execution.

\paragraph{Kernel-level runtime breakdown.}
Table~\ref{tab:kernel-breakdown} reports the time spent in major GPU kernel components. Model computation accounts for the largest fraction of GPU execution time, including 33.3\% in decode-only batches and 35.4\% in batches containing both decoding and refresh. Sparse attention and Quest page selection account for 16.8\% and 3.4\%, respectively. Entropy evaluation and trigger checking account for only 0.25\% of GPU kernel time.

\begin{table}[t]
    \centering
    \small
    \setlength{\tabcolsep}{8pt}
    \begin{tabular}{lr}
        \toprule
        Component & GPU Time (\%) \\
        \midrule
        Model computation (decode-only batches) & 33.3 \\
        Model computation (mixed decode/refresh batches) & 35.4 \\
        Sparse attention & 16.8 \\
        Sampling & 4.1 \\
        Dense attention (mixed batches) & 3.9 \\
        Quest page selection & 3.4 \\
        KV-cache writes & 0.7 \\
        Standalone refresh forward & 0.3 \\
        Entropy evaluation and trigger checking & 0.25 \\
        \bottomrule
    \end{tabular}
    \caption{Kernel-level runtime breakdown of \method{} on Qwen3-8B/Math87. Percentages are relative to total GPU kernel execution time.}
    \label{tab:kernel-breakdown}
\end{table}

\paragraph{Incremental refresh overhead.}
The kernel-level breakdown does not directly measure the overhead introduced by dense KV refresh, since decoding and refresh can execute within the same batch. In particular, model computation in mixed batches cannot be attributed exclusively to either operation.

To estimate the incremental cost of refresh, we stratify 220 scheduler steps by execution phase and batch type. For each sampled batch, we compare forward execution with and without refresh rows under identical initial KV states, and weight the results according to the number of scheduler steps in each stratum. This gives an estimated incremental refresh-forward cost of 126.55 seconds, corresponding to approximately 6.0\% of total wall-clock time. Entropy evaluation and trigger checking account for 15.33 seconds, or approximately 0.7\% of wall-clock time.

These estimates characterize the additional cost of individual components along the observed execution trajectories, rather than the end-to-end speedup that would result from disabling them.

\paragraph{Scheduling overhead.}
We additionally examine whether irregular refresh scheduling introduces GPU idle time. Across two GPU allocations, idle time accounts for 4.6\% and 10.2\% of wall-clock time, respectively. Since our implementation uses a single CUDA stream, these measurements provide an upper bound on idle time attributable to host-side scheduling, planning, and kernel-launch overhead. We observe no consistent increase in per-step idle time when refreshes are present.

\end{document}